\documentclass[10pt,twocolumn,letterpaper]{article}

\usepackage[pagenumbers]{cvpr} 

\usepackage{graphicx}
\usepackage{amsmath}
\usepackage{amssymb}
\usepackage[accsupp]{axessibility}  
\usepackage{bm}
\usepackage{booktabs}
\usepackage{array,multirow}
\usepackage{float}
\usepackage{expl3, xparse, siunitx}
\ExplSyntaxOn
\NewDocumentCommand { \calcnum } { O{} m }
  { \num [  round-mode=places , round-precision=4 , group-separator={,}, group-minimum-digits=6, #1] { \fp_to_decimal:n {#2} } }
\ExplSyntaxOff
\usepackage[table]{xcolor}
\definecolor{lightgray}{gray}{0.9}
\definecolor{lightblue}{rgb}{0.93,0.95,1.0}
\definecolor{lighterblue}{rgb}{0.96,0.98,1.0}
\definecolor{lightgreen}{rgb}{0.93,1.0,0.95}
\definecolor{lightergreen}{rgb}{0.96,1.0,0.98}
\definecolor{darkgreen}{rgb}{0.0,0.6,0.0}
\definecolor{blue}{rgb}{0, 0, 1}

\newcolumntype{j}{>{\columncolor{lighterblue}}c}
\newcolumntype{k}{>{\columncolor{lightergreen}}c}

\newcommand{\minisection}[1]{\vspace{1mm}\noindent{\textbf{#1}.}}

\newenvironment{tight_itemize}{
\begin{itemize}
  \setlength{\topsep}{0pt}
  \setlength{\itemsep}{2pt}
  \setlength{\parskip}{0pt}
  \setlength{\parsep}{0pt}
}{\end{itemize}}

\usepackage[pagebackref,breaklinks,colorlinks]{hyperref}

\usepackage[capitalize]{cleveref}
\crefname{section}{Sec.}{Secs.}
\Crefname{section}{Section}{Sections}
\Crefname{table}{Table}{Tables}
\crefname{table}{Tab.}{Tabs.}

\DeclareMathOperator*{\argmin}{\arg\!\min}

\begin{document}

\title{Few-shot Learning with Noisy Labels}

\author{Kevin J Liang$^{1}$ \quad Samrudhdhi B. Rangrej$^{2}$ \quad Vladan Petrovic$^{1}$ \quad Tal Hassner$^{1}$ \\
$^{1}$Facebook AI Research \quad$^{2}$McGill University \\
{\tt\small kevinjliang@fb.com}}
\maketitle

\graphicspath{{Figures/}}

\begin{abstract}
Few-shot learning (FSL) methods typically assume clean support sets with accurately labeled samples when training on novel classes. This assumption can often be unrealistic: support sets, no matter how small, can still include mislabeled samples. Robustness to label noise is therefore essential for FSL methods to be practical, but this problem surprisingly remains largely unexplored. To address mislabeled samples in FSL settings, we make several technical contributions. (1) We offer simple, yet effective, feature aggregation methods, improving the prototypes used by ProtoNet, a popular FSL technique. (2) We describe a novel Transformer model for Noisy Few-Shot Learning (TraNFS). TraNFS leverages a transformer's attention mechanism to weigh mislabeled versus correct samples. (3) Finally, we extensively test these methods on noisy versions of MiniImageNet and TieredImageNet. Our results show that TraNFS is on-par with leading FSL methods on clean support sets, yet outperforms them, by far, in the presence of label noise.
\end{abstract}

\vspace{-3mm}
\section{Introduction}
\label{sec:intro}
\vspace{-2mm}
Modern few-shot learning (FSL) methods aim to learn classifiers for novel classes from only a handful of examples. These methods, however, generally assume that the few {\em support set} samples used for training were carefully selected to represent their class. Unfortunately, real-world settings rarely offer such guarantees. In fact, even carefully annotated and curated datasets often contain mislabeled samples~\cite{tsipras2020imagenet, beyer2020we, masi2019face, yang2020object, northcutt2021pervasive}, due to automated weakly supervised annotation, ambiguity, or even human error. 

Whereas there are plenty of methods designed for learning with noise in many-shot supervised settings~\cite{angluin1988learning, natarajan2013learning, li2017learning, wang2018iterative, jiang2018mentornet, han2018co}, noise in few-shot settings remains largely unexplored. This dearth is surprising considering the utility of FSL methods in settings where human supervision cannot easily be provided: \eg in fully automated systems which learn many novel classes~\cite{davidson2010youtube, toledo2015correcting, kaufman2019balancing, ye2020purifynet, yin2022sylph}, making human curation of the labels of every support set, unrealistic. 

\begin{figure}
\centering
\includegraphics[width=0.8\columnwidth]{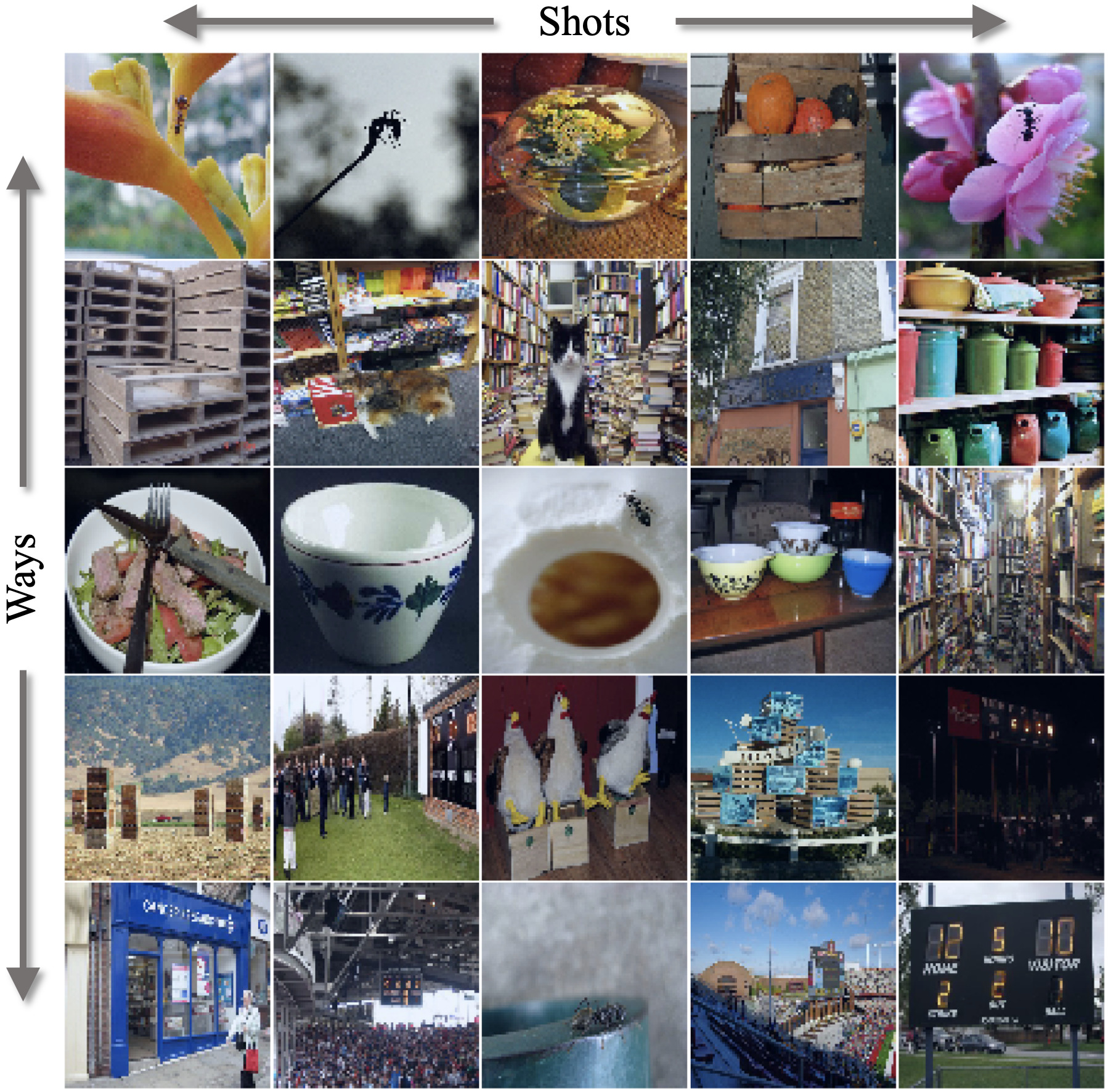}
\vspace{-2mm}
\caption{{\bf Few-shot learning with mislabeled samples.} A 5-shot, 5-way support set of MiniImageNet~\cite{vinyals2016matching} images. Rows show support set samples of each novel class. Two samples in each row were mislabeled by symmetric label flips (Sec.~\ref{sec:setup}). Can you spot which ones? See Appx.~\ref{apx:nfsl_ex} for answers and more examples.}
\label{fig:qualitative}
\vspace{-4mm}
\end{figure}

Fig.~\ref{fig:qualitative} shows the challenge of learning from few, possibly mislabeled, examples. It presents a sample 5-shot, 5-way support set from MiniImageNet~\cite{vinyals2016matching}. 
Each row includes the support set training images of one of the five classes. 
Two of the samples in each row are mislabeled with symmetric label noise (Sec.~\ref{sec:setup}). With so few examples, spotting mislabeled images can be difficult, even for humans with considerable prior knowledge, which FSL methods lack.

\begin{figure}
\centering
\includegraphics[width=0.86\columnwidth]{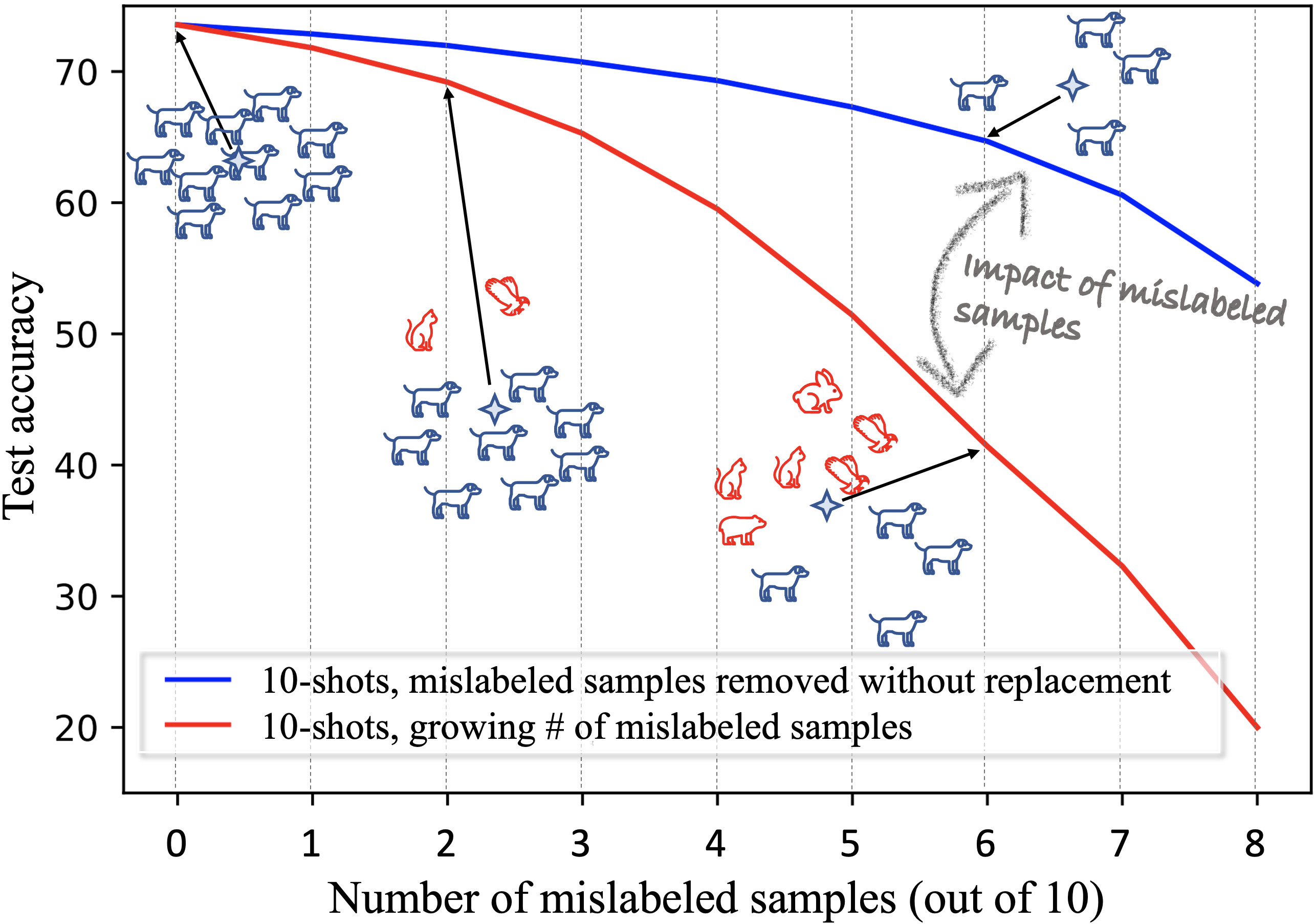}
\vspace{-2mm}
\caption{{\bf Number of mislabeled samples versus few-shot learning accuracy.} Accuracy for 10-shot, 5-way classification on MiniImageNet~\cite{vinyals2016matching} reported for a vanilla ProtoNet~\cite{snell2017prototypical}.
The animals represent support set embeddings, with the mean prototype (star) being pulled out of the clean class (dog) distribution with increasing mislabeled samples.
{{\textcolor{blue}{Blue:}}} Accuracy if mislabeled samples are known and ignored. {{\textcolor{red}{Red:}}} Accuracy when using full support sets without removing mislabeled samples. The gap between these two curves reflects the vulnerability of few-shot learning to label noise.}
\label{fig:proto_noisy}
\vspace{-4mm}
\end{figure}

As we later demonstrate empirically, FSL methods are especially vulnerable to such label noise. When training from few samples, each sample represents a significant contribution to the final decision boundary. Thus, even a single noisy example can be destructive to the model's accuracy. We illustrate this observation in Fig.~\ref{fig:proto_noisy}, which reports the performance of ProtoNet~\cite{snell2017prototypical}, a popular FSL method, on MiniImageNet with noisy labels. ProtoNet averages the convolutional features of each class's support set into class {\em prototypes}. Queries are then classified by the class of their nearest neighbor prototype. Fig.~\ref{fig:proto_noisy} shows the effect of increasing the number of mislabeled samples, compared with a model trained after mislabeled samples were removed (\ie, smaller, but cleaner, support sets). The widening gap between the two curves reflects the degradation of accuracy when mislabeled samples are not accounted for. 

We address the vulnerability of FSL methods to label noise by making a number of technical innovations. We begin by exploring simple, yet effective alternatives to the design of ProtoNet~\cite{snell2017prototypical}. Specifically, we replace the mean operator, used by ProtoNet for aggregating support set features, with more robust methods. We evaluate an unweighted option, the median, and options which weigh support set samples based on feature similarities. We show that these changes already improve robustness to label noise.

We then introduce our \textbf{Tra}nsformer model for \textbf{N}oisy \textbf{F}ew-\textbf{S}hot Learning (TraNFS). Unlike previous methods, TraNFS {\em learns} to aggregate support samples into class representations. The transformer architecture offers a natural means for processing variable numbers of shots and ways with permutation invariance. Robustness to label noise is achieved by leveraging a modified version of the transformer's self-attention mechanism~\cite{vaswani2017attention}. This modified self-attention used by TraNFS compares support set samples and downweights samples considered likely to be mislabeled. 

We test our proposed methods extensively on versions of MiniImageNet~\cite{vinyals2016matching} and TieredImageNet~\cite{ren2018meta} with three methods of adding label noise. 
Our results show that the proposed TraNFS (and even the simpler modifications of ProtoNet) surpass popular FSL methods by wide margins in the presence of label noise, while offering comparable performance in the absence of label noise.  

To summarize, we make the following contributions.
\vspace{-2mm}
\begin{tight_itemize}
    \item We propose median and similarity weighting as simple yet effective substitutes to ProtoNet's mean prototypes.
    \item We present TraNFS, a novel transformer model adapted to FSL with noisy labels.
    \item We extensively benchmark many popular FSL methods on three types of support set noise pollution: symmetric, paired, and outlier.
\end{tight_itemize}
\vspace{-2mm}
Our code can be found at \url{https://github.com/facebookresearch/noisy_few_shot}.
\vspace{-1mm}

\section{Related work}
\label{sec:relwork}
\vspace{-2mm}
\minisection{Few-shot learning}
The field of FSL methods is vast; we refer to surveys for comprehensive overviews~\cite{wang2020generalizing,bendre2020learning}. 

Metric-based methods classify query samples based on their similarity to each class's support examples, learning a transferable embedding space for which such comparisons can be made. Metrics such as cosine similarity~\cite{vinyals2016matching}, Euclidean distance~\cite{snell2017prototypical}, Mahalanobis distance~\cite{bateni2020improved}, and Earth Mover’s Distance (EMD)~\cite{zhang2020deepemd} have been shown to be effective. RelationNet~\cite{sung2018learning} and Satorras \etal~\cite{satorras2018few} used convolutional and graph neural networks, respectively, to learn a similarity metric. TADAM~\cite{oreshkin2018tadam}, FEAT \cite{ye2020few}, and TAFE-Net~\cite{wang2019tafe} proposed task-specific adaptation of embeddings. 
CrossTransformers~\cite{doersch2020crosstransformers} used attention for spatially-aware similarity between local features.

Optimization-based methods fine-tune model parameters on few support examples. MAML~\cite{finn2017model, antoniou2018train} learned model parameter initializations that allow fast fine-tuning on few samples. REPTILE~\cite{nichol2018first} simplified MAML with a first-order formulation. 
MetaNet~\cite{munkhdalai2017meta} introduced fast and slow weights for fast parameterization and rapid generalization. 
Bertinetto \etal~\cite{bertinetto2018meta} and MetaOptNet~\cite{lee2019meta} presented closed form solutions and differentiable solvers for task-dependent Ridge Regression, Logistic Regression (LR), and Support Vector Machines (SVMs). 
Tian \etal~\cite{tian2020rethinking} showed learning of generalizable feature embeddings used to train linear classifiers on novel tasks.

\minisection{Noisy labels and outliers}
Methods for learning noise transition matrices are common~\cite{liu2015classification, yao2020dual, zhang2021learning}. Estimating a noise transition matrix from a handful of potentially mislabeled samples, however, is an ill-posed problem. Other methods~\cite{jiang2018mentornet,han2018co,yu2019does} leverage deep neural networks' tendency to learn easier (and thus likely correctly labeled) samples first~\cite{arpit2017closer, toneva2018empirical} to select reliable samples to learn from, but such behavior cannot be relied upon when only a few samples are available.
Deep out-of-distribution (OOD) detection is also extensively explored~\cite{devries2018learning, Vyas_2018_ECCV, ren2019likelihood, bibas2021single}, but these methods typically focus on identifying test-time outliers that are out-of-distribution relative to the training set. In FSL, disjoint base and novel sets mean that \textit{all} meta-test samples are considered OOD, including correctly labeled support set samples.
Finally, there are several works that take a meta-learning approach to learning noisy labels~\cite{li2019learning, wang2020training, zheng2021meta}, but these methods typically assume known label spaces with abundant data (\ie many-shots), rather than our few-shot setting. With only few training samples, these methods fail.

\minisection{Robust FSL}
With few previous works, noisy labels have largely been ignored by FSL methods. 
RNNP~\cite{mazumder2021rnnp} combined data augmentation with repeated applications of $k$-means to produce refined prototypes, but such unsupervised clustering implicitly assumes that noisy data is from one of the support set classes. RapNets~\cite{lu2020robust} proposed a BiLSTM-based attentive module to overcome representation or label noise.
Alternatively, RW-MAML~\cite{killamsetty2020reweighted} learned to weigh support samples by extending MAML to bi-bi-level optimization, but it considers the less realistic setting of mixing in OOD tasks during metatraining rather than noisy few-shot meta-test. 
Finally, robustness of meta-learners to adversarial attacks have also been considered~\cite{goldblum2020adversarially}. 

\section{Preliminaries}
\label{sec:prelim}
\vspace{-2mm}
FSL classification tasks are often referred to as $K$-shot $N$-way, where $N$ is the number of classes being learned and $K$ is the number of labeled samples per class to learn from. These $KN$ samples $S = \{x_1^{(1)}, x_2^{(1)}, ..., x_{K-1}^{(N)}, x_K^{(N)}\}$ are often referred to as the {\em support set}. After training, unlabeled queries are classified into one of these $N$ classes. To produce an effective classifier of novel classes $C^n$ from few samples, FSL models typically use knowledge transfer, leveraging a dataset of {\em base classes} $C^b$ with ample labeled data. It is commonly assumed that the classes in $C^n$ are unknown in advance and thus absent from $C^b$ (\ie  $C^b \cap C^n = \emptyset$). Recent FSL methods often adopt a meta-learning paradigm, simulating the desired inference-time behavior by meta-training the model with many episodes of $K$-shot, $N$-way tasks, optimizing for accuracy on $Q$ query samples from each of the episode's $N$ classes.

\minisection{ProtoNets} One such relevant FSL method is Prototypical Networks (ProtoNets)~\cite{snell2017prototypical}. 
ProtoNets use a convolutional feature extractor $\mathcal F$ to convert each sample in the support set to an embedding $h_i^{(c)} = \mathcal F(x_i^{(c)}) \in \mathbb{R}^D$. These embeddings are then aggregated into $N$ class prototypes $p^{(c)}$ using a simple mean of the embeddings for each class $c$:
\begin{equation}
    p^{(c)} = \frac{1}{K} \sum_i \mathcal{F}(x_i^{(c)}).
\end{equation}
A query sample, $x_q$, is then classified based on the nearest prototype in embedding space: 
\begin{equation}
    y = \argmin_c d(\mathcal{F}(x_q), p^{(c)}).
\end{equation}
Despite its simplicity, ProtoNets remain a strong baseline, and its easy implementation makes it compelling for real-world use cases at scale. Using mean to aggregate embeddings, however, implies sensitivity to mislabeled samples, especially when only few samples are provided. Indeed, as we show in Fig.~\ref{fig:proto_noisy}, incorrectly labeled samples can easily degrade accuracy of the resulting classifiers. This is a symptom of the prototypes being {\em pulled away} from the class's true (unknown) mean by mislabeled samples. 

\vspace{-2mm}
\section{Static alternatives to the mean}
\label{sec:static_agg}
\vspace{-2mm}
Using mean as proposed by ProtoNet~\cite{snell2017prototypical} to aggregate features is not the only way to combine embeddings into prototypes: other aggregation methods may be better suited when mislabeled samples are expected. 
We begin by exploring simple alternatives to the mean, intended to make prototypes more robust to mislabeled samples while maintaining accuracy if all labels are correct.

\subsection{Spatial median prototypes}
\label{sec:median}
\vspace{-2mm}
The median is a natural alternative to the mean in noisy settings. While order statistics like the median are well defined for scalars, this is not the case for vectors. For scalars, there is a connection between various distribution statistics (\eg mean, median, mode) and minimization of the appropriate loss functions~\cite{BarronCVPR2019RobustLoss}. For example, empirical mean minimizes total squared error between the mean and values in the set. Similarly, empirical median minimizes total absolute error between the median and the set, so finding a median is equivalent to minimizing the total absolute error.

This minimization generalizes well to higher dimensional spaces. We thus define a cost function to be the sum of distances to embedding vectors $h_i^{(c)}, i\in\{1, 2, ..., K\}$, in the set for each class, $c$, and find the median vector $p^{(c)}$ minimizing this cost. For brevity, we drop the class index $c$ in derivations that follow. To make the loss differentiable at all points, we work with a smooth version of the loss, usually referred to as the {\em pseudo-Huber} loss:
\begin{equation}
    \mathcal L(p) = \sum_{i = 1}^K \left(\sqrt{||p - h_i||_2^2 + \epsilon^2} - \epsilon\right),
\end{equation}
\noindent where $K$ is the number of vectors in the set, $||\cdot||_2$ is an $L^2$ vector norm, and $\epsilon$ is a small constant.

No closed-form solution for this minimization problem exists, so we use Newton's method for an iterative solution:
\begin{equation}
    p(t + 1) = p(t) - \mathcal H^{-1}(p(t)) \cdot \nabla \mathcal L(p(t)).
\end{equation}
\noindent We find the gradient, $\nabla \mathcal L(p)$, and the Hessian matrix, $\mathcal H(p)$, using matrix calculus with numerator layout as,
\begin{equation}
    \nabla \mathcal L(p) = \sum_{i = 1}^K \frac{p - h_i}{\sqrt{||p - h_i||_2^2 + \epsilon^2}},
\end{equation}
\begin{equation}
    \mathcal H(p) = \left(\sum_{i = 1}^K \frac{1}{\sqrt{||p - h_i||_2^2 + \epsilon^2}}\right)I_{D \times D} - UU^T,
\end{equation}
\noindent where $D$ is the dimension of the vector space, $I_{D \times D}$ is the identity matrix, and $U = [u_1, u_2, \dots, u_K]$ is a $D \times K$ matrix formed by stacking vectors $u_i = \frac{p - h_i}{(||p - h_i||_2^2 + \epsilon^2)^\frac{3}{4}}$. As an approximation, we can neglect the second, non-diagonal term in the Hessian, in which case the iteration becomes:
\begin{equation}\label{eq:medupdate}
    p(t + 1) = p(t) - \frac{\sum_{i = 1}^K \frac{p(t) - h_i}{\sqrt{||p(t) - h_i||_2^2 + \epsilon^2}}}{\sum_{i = 1}^K \frac{1}{\sqrt{||p(t) - h_i||_2^2 + \epsilon^2}}}.
\end{equation}

\noindent Note that the choice of pseudo-Huber loss with a small constant $\epsilon$ avoids division by zero even when the median estimate falls exactly at one of the vectors in the support set.
See Appx.~\ref{apx:median} for additional comments on complexity.

\begin{figure*}[t]
    \centering
    \includegraphics[width=0.8\textwidth]{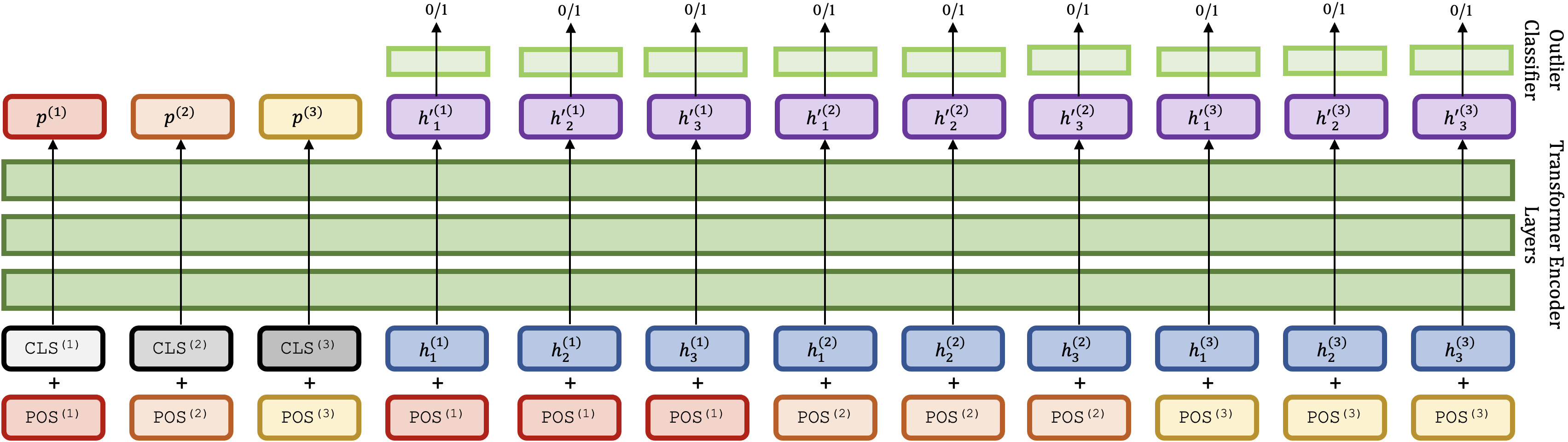}
    \vspace{-2mm}
    \caption{{\bf Visualization of our proposed TraNFS architecture}, for a 3-shot 3-way support set example input / output sequence.}
    \label{fig:tranfs}
    \vspace{-4mm}
\end{figure*}

\subsection{Similarity weighted prototypes}
\label{sec:sim_proto}
\vspace{-1mm}
ProtoNet-style mean aggregation uniformly weighs all shots of a class's support set. A clear extension to this scheme is a non-uniform weighting which suppresses outliers and amplifies clean samples. Of course, if we knew which samples were mislabeled, we could remove them from the support set, but this information is typically unavailable. Instead, we can try to identify mislabeled samples based on how the support set is arranged in feature space.

Specifically, we assume that a well-trained feature extractor $\mathcal{F}$ embeds correctly labeled samples close to one another~\cite{khosla2020supervised}, thus on average being closer in the induced metric space than any mislabeled samples. This intuition implies that the subset of correctly labeled samples is larger than any subset of mislabeled samples from a single, unrelated class. This assumption, however, is typical to many robust estimators, including, \eg, the Random Sample Consensus (RANSAC)~\cite{fischler1981random}. Building on this assumption, we offer the following similarity measures.

\minisection{Squared Euclidean distance}
This distance is the same one minimized by ProtoNets and thus a natural choice for measuring distances when attempting to identify mislabel samples. We compute the similarity score as:
\begin{equation}
    a_i^{(c)} = - \frac{1}{K-1} \sum_{i \neq j} || h_i^{(c)} - h_j^{(c)} ||^2_2.
\end{equation}
\noindent Lower distance implies being closer to other support samples, so we negate the average distance for the final score.

\minisection{Absolute distance} The $L^2$ norm can heavily penalize large distances in few feature dimensions: a large difference in only few dimensions may result in a large distance between the features, even if they share similar values in all other dimensions. We thus also consider $L^1$:
\begin{equation}
    a_i^{(c)} = - \frac{1}{K-1} \sum_{i \neq j} | h_i^{(c)} - h_j^{(c)} |.
\end{equation}
As with the Euclidean distance, we use a factor of $-1$ so that lower distances produce higher scores.

\minisection{Cosine similarity} While not a proper distance metric, the cosine angle between two features is a common measure of feature similarity in the few-shot literature~\cite{chen2019closer}.
\begin{equation}
    a_i^{(c)} = \frac{1}{K-1} \sum_{i \neq j} \frac{h_i^{(c)} \cdot h_j^{(c)}}{||h_i^{(c)}||~ ||h_j^{(c)}||}.
\end{equation}
\noindent As the inputs are normalized, cosine similarity is less sensitive to the magnitude of the embeddings.

\minisection{Aggregating features with weighted similarity} Once we obtain the average distance of each feature to others in the same support set, using one of the scores above, we produce an aggregated prototype by weighting the support samples using these scores, normalizing the result with a softmax.
\begin{align}
    w_i^{(c)} &= \frac{\mathrm{exp}(a_i^{(c)}/T)}{\sum_j \mathrm{exp}(a_j^{(c)}/T)}, \\
    p^{(c)} &= \sum_i w_i^{(c)} \mathcal{F}(x_i^{(c)}),
\end{align}
\noindent where $T$ is a temperature term controlling the diffuseness of the softmax. As $T\rightarrow 0$, this method picks the support sample with minimum distance to the other samples as a class prototype, while as $T\rightarrow \infty$, this reduces to the mean (\ie ProtoNets~\cite{snell2017prototypical}). We choose soft-weighting of support samples, rather than top-$k$ selection or a hard threshold, as the latter two require either knowing the number of noisy samples or threshold tuning, which may vary depending on the class or support sample distribution.

\section{Learning a prototype aggregator}
\vspace{-2mm}
The aggregation methods discussed in Sec.~\ref{sec:static_agg}, weighted or otherwise, are hard coded: They do not adjust to differences in support set feature distributions. We hypothesize that a learned mechanism that compares support set embeddings for similarity {\em and then refines them}, can potentially outperform these static methods. Crucially, in typical FSL settings, the number and order of support samples and classes are arbitrary. Thus, any learned alternative to the methods of Sec.~\ref{sec:static_agg} must process arbitrary numbers of shots or ways while remaining permutationally invariant to both.

\subsection{A transformer model for noisy FSL}
\vspace{-2mm}
Given these requirements, we propose a \textbf{Tra}nsformer model for \textbf{N}oisy \textbf{F}ew-\textbf{S}hot Learning (TraNFS) (Fig.~\ref{fig:tranfs}). Transformers are designed to process sequences of arbitrary length while offering permutation invariance. Importantly, we note that a transformer's self-attention mechanism~\cite{vaswani2017attention} can be leveraged to compute similarities between support set samples and naturally weigh them when aggregating them into prototypes. 
To this end, we concatenate the convolutional features of a support set's samples to form an input sequence $\bm h = [h_1^{(1)}, h_2^{(1)}, ..., h_{K-1}^{(N)}, h_K^{(N)}]$ to the transformer, $\mathcal T$.
We then make the following adaptations to the transformer to enable it to process a typical FSL support set.

\minisection{Class token} Partly inspired by BERT~\cite{devlin2019bert}, we use a set of classification tokens $\texttt{CLS}^{(c)}, c \in \{1, ..., N\}$ to denote the positions representing the prototype for each class and concatenate $[\texttt{CLS}^{(1)},..., \texttt{CLS}^{(N)}]$ to the support set embedding sequence $\bm h$.
By taking the output at the position of $\texttt{CLS}^{(c)}$ to be the prototype $p^{(c)}$ for class $c$, we motivate the transformer to learn to aggregate the information in all support set samples into this position.
There are multiple choices for instantiating $\texttt{CLS}^{(c)}$, including as a random constant, the mean of the support set embeddings for class $c$ (\ie a mean prototype), or a learnable embedding.
We report comparisons of these variations in the supplemental material.

\minisection{Positional encoding}
Shot and class order are both typically arbitrary in FSL and should therefore not be encoded. Still, we require some means of informing the transformer the class identity of each support sample. Vaswani \etal~\cite{vaswani2017attention} utilized a sinusoidal positional encoding added to the input sequence to indicate word order. We repurpose this mechanism and use it to encode the class $c$ associated with each position in the input sequence.
Specifically, we create $N$ $D$-dimensional embeddings corresponding to special tokens $\texttt{POS}^{(c)}, c \in \{1, ..., N\}$ and add each $\texttt{POS}^{(c)}$ to all support sample embeddings $h_i^{(c)}$ and the class tokens $\texttt{CLS}^{(c)}$, as seen in Fig.~\ref{fig:tranfs}.
With the positional embeddings added to the input sequence, the transformer can learn to attend to the positional encoding to associate support set embeddings and the prototype position of each class together.

\vspace{-1mm}
\subsection{Optimization}
\label{sec:method_opt}
\vspace{-2mm}
We meta-train TraNFS to minimize the standard ProtoNet loss. Logits are computed as the negative distance $d$ between prototypes predicted by the model at the \texttt{CLS} token positions and the embedded query sample $\mathcal F(x_q)$:
\begin{equation}
    \mathcal L_{\mathrm{xent}} =  -\sum_{c=1}^N y_q \cdot \log\left(\frac{\mathrm{exp}\left(-d\left(p^{(c)}, \mathcal F(x_q)\right)\right)}{\sum_{c'} \mathrm{exp}\left(-d\left(p^{(c')}, \mathcal F(x_q)\right)\right)} \right)
\end{equation}
\noindent where $\cdot$ is the dot product and $y_q$ is the one-hot ground truth label of query $x_q$.

When meta-training TraNFS, we found it essential to expose the model to support sets with noisy samples (Sec.~\ref{sec:noise_prop}). We do this by artificially introducing label noise to the support set, using label $o_i^{(c)} \in \{0, 1\}$ to track the positions of the noisy samples. This step ensures that the transformer learns a noise rejection mechanism. Without noisy samples, the transformer is not motivated to learn anything beyond recreating ProtoNet by averaging support samples. 

\begin{table*}[t!]
\centering
\caption{
{\bf Few-shot with symmetric label swap noise}. 5-way 5-shot Acc. $\pm$ 95\% CI on {\textcolor{blue}{MiniImageNet}}~\cite{vinyals2016matching}, {\textcolor{darkgreen}{TieredImageNet}}~\cite{ren2018meta}.
Our TraNSF is comparable to existing methods at $0\%$ noise, with a growing gap in its favor as noise levels increase.
Best viewed in color.
}
\vspace{-3mm}
\setlength{\tabcolsep}{6pt}
\def\arraystretch{1.05}
\resizebox{0.95\textwidth}{!}{
\begin{tabular}{c c || j k j k j k j k} 
    \toprule[1.2pt]
& Model \textbackslash ~ Noise Proportion & \multicolumn{2}{c}{$0\%$}  & \multicolumn{2}{c}{$20\%$} & \multicolumn{2}{c}{$40\%$}  & \multicolumn{2}{c}{$60\%$} \\
\toprule[0.8pt]		
& Oracle             & 68.18 $\pm$ 0.16 & 71.42 $\pm$ 0.18 & 66.08 $\pm$ 0.17 & 69.19 $\pm$ 0.19 & 62.60 $\pm$ 0.17 & 66.14 $\pm$ 0.20 & 56.89 $\pm$ 0.18 & 60.39 $\pm$ 0.21 \\
\hline
{\multirow{9}{*}{\rotatebox[origin=c]{90}{Baselines}}} &
Nearest $k=1$      & 55.91 $\pm$ 0.17 & 58.81 $\pm$ 0.20 & 47.27 $\pm$ 0.18 & 49.48 $\pm$ 0.19 & 38.68 $\pm$ 0.18 & 40.25 $\pm$ 0.19 & 29.20 $\pm$ 0.16 & 29.84 $\pm$ 0.17 \\
& Nearest $k=3$      & 55.29 $\pm$ 0.18 & 58.44 $\pm$ 0.20 & 48.43 $\pm$ 0.17 & 51.11 $\pm$ 0.19 & 39.14 $\pm$ 0.17 & 41.09 $\pm$ 0.18 & 29.66 $\pm$ 0.15 & 30.69 $\pm$ 0.15 \\
& Nearest $k=5$      & 56.15 $\pm$ 0.18 & 59.22 $\pm$ 0.20 & 50.92 $\pm$ 0.17 & 53.75 $\pm$ 0.19 & 42.12 $\pm$ 0.17 & 44.14 $\pm$ 0.19 & 32.62 $\pm$ 0.16 & 33.99 $\pm$ 0.17 \\
& Linear Classifier  & 66.65 $\pm$ 0.16 & 69.89 $\pm$ 0.18 & 58.41 $\pm$ 0.17 & 61.96 $\pm$ 0.19 & 47.23 $\pm$ 0.17 & 50.08 $\pm$ 0.19 & 34.04 $\pm$ 0.16 & 35.75 $\pm$ 0.17 \\
& Matching Networks~\cite{vinyals2016matching}  & 62.16 $\pm$ 0.17 & 64.92 $\pm$ 0.19 & 56.21 $\pm$ 0.18 & 59.20 $\pm$ 0.20 & 46.18 $\pm$ 0.18 & 49.12 $\pm$ 0.20 & 34.66 $\pm$ 0.18 & 36.80 $\pm$ 0.19 \\
& MAML~\cite{finn2017model} & 63.25 $\pm$ 0.18 & 63.96 $\pm$ 0.19 & 53.28 $\pm$ 0.18 & 54.62 $\pm$ 0.19 & 42.58 $\pm$ 0.18 & 43.71 $\pm$ 0.19 & 31.01 $\pm$ 0.17 & 31.74 $\pm$ 0.17 \\
& Vanilla ProtoNet~\cite{snell2017prototypical} & 68.27 $\pm$ 0.16 & 71.36 $\pm$ 0.18 & 62.43 $\pm$ 0.17 & 66.15 $\pm$ 0.19 & 51.41 $\pm$ 0.19 & 55.05 $\pm$ 0.21 & 38.33 $\pm$ 0.19 & 40.61 $\pm$ 0.21 \\
& Baseline++~\cite{chen2019closer} & 67.91 $\pm$ 0.16 & 71.24 $\pm$ 0.18 & 61.87 $\pm$ 0.17 & 65.58 $\pm$ 0.19 & 51.87 $\pm$ 0.18 & 55.00 $\pm$ 0.20 & 38.36 $\pm$ 0.19 & 40.19 $\pm$ 0.20 \\
& RNNP~\cite{mazumder2021rnnp} & 68.38 $\pm$ 0.16 & 71.36 $\pm$ 0.18 & 62.43 $\pm$ 0.17 & 65.95 $\pm$ 0.19 & 51.62 $\pm$ 0.19 & 54.86 $\pm$ 0.21 & 38.45 $\pm$ 0.19 & 40.63 $\pm$ 0.21 \\
\hline
{\multirow{6}{*}{\rotatebox[origin=c]{90}{Ours}}}
& Median             & 68.45 $\pm$ 0.16 & 71.28 $\pm$ 0.18 & 63.19 $\pm$ 0.17 & 66.65 $\pm$ 0.20 & 51.86 $\pm$ 0.19 & 55.09 $\pm$ 0.21 & 39.32 $\pm$ 0.19 & 41.94 $\pm$ 0.21 \\
& Absolute  & 68.24 $\pm$ 0.16 & 71.27 $\pm$ 0.18 & 63.46 $\pm$ 0.17 & 66.87 $\pm$ 0.20 & 52.06 $\pm$ 0.20 & 55.26 $\pm$ 0.22 & 39.78 $\pm$ 0.20 & 42.54 $\pm$ 0.22 \\
& Euclidean & 68.32 $\pm$ 0.16 & \textbf{71.48 $\pm$ 0.18} & 63.02 $\pm$ 0.17 & 66.69 $\pm$ 0.19 & 52.09 $\pm$ 0.19 & 55.62 $\pm$ 0.21 & 39.33 $\pm$ 0.20 & 41.75 $\pm$ 0.21 \\
& Cosine      & 68.20 $\pm$ 0.16 & 70.59 $\pm$ 0.18 & 63.46 $\pm$ 0.17 & 66.62 $\pm$ 0.20 & 52.42 $\pm$ 0.20 & 55.78 $\pm$ 0.22 & 39.90 $\pm$ 0.20 & 42.56 $\pm$ 0.22 \\
\cmidrule[0.8pt]{2-10}
& TraNFS-$2$	       & 68.29 $\pm$ 0.17 & 70.92 $\pm$ 0.19 & 64.74 $\pm$ 0.18 & 67.33 $\pm$ 0.21 & 56.14 $\pm$ 0.21 & 58.76 $\pm$ 0.23 & 42.24 $\pm$ 0.23 & 44.17 $\pm$ 0.25 \\
& TraNFS-$3$         & \textbf{68.53 $\pm$ 0.17} & 71.17 $\pm$ 0.19 & \textbf{65.08 $\pm$ 0.18} & \textbf{67.67 $\pm$ 0.20} & \textbf{56.65 $\pm$ 0.21} & \textbf{58.88 $\pm$ 0.23} & \textbf{42.60 $\pm$ 0.24} & \textbf{44.21 $\pm$ 0.25} \\
\bottomrule[1.2pt]
\end{tabular}
}
\vspace{-3mm}
\label{tab:sym_swap_5w5k}
\end{table*}

\minisection{Clean prototype loss} Besides optimizing the position of predicted prototypes relative to the meta-training query samples, we also encourage the predicted prototype for each class to be close to a {\em clean} prototype, $\hat{p}^{(c)}$, aggregated from correctly labeled samples in the support set:
\begin{equation}
    \hat{p}^{(c)} = \frac{1}{K - \sum_i o_i^{(c)}} \sum_{i} \mathbf{1}[o_i^{(c)}=0] \mathcal F(x_i^{(c)}), 
\end{equation}
\begin{equation} \label{eq:clean_proto}
\mathcal L_{\mathrm{clean}} = \frac{1}{N} \sum_c ||p^{(c)}- \hat{p}^{(c)}||^2_2.
\end{equation}
\noindent We choose mean squared error here, but other alternatives such as negative cosine similarity are also viable.

\minisection{Binary outlier classification loss} The ProtoNet and clean prototype losses described above both implicitly encourage identification of noisy samples. We found it helpful to also explicitly train the model to classify support set samples as either mislabeled or not.

We instantiate the binary classifier as a fully connected layer $\mathcal B$ applied to the transformer's output at positions corresponding to the support set samples. We share weights for $\mathcal B$ across all such positions, with loss term:
\begin{align} \label{eq:bin_cls}
    \mathcal L_{\mathrm{bin}} = -\frac{1}{KN} & \sum_{i,c}  o_i^{(c)} \log \sigma(\mathcal B({h'}_i^{(c)})), \\ \nonumber
    & + (1- o_i^{(c)}) \log \left(1 - \sigma(\mathcal B({h'}_i^{(c)}))\right),
\end{align}
\noindent where $\sigma$ is the sigmoid function and ${h'}_i^{(c)}$ is the transformer output corresponding to $h_i^{(c)}$.

Our final optimization objective combines the three losses described above: 
\begin{equation}\label{eq:loss}
    \mathcal L = \mathcal L_{\mathrm{xent}} + \lambda_c \mathcal  L_{\mathrm{clean}} + \lambda_b \mathcal L_{\mathrm{bin}}, 
\end{equation}
\noindent where $\lambda_c$ and $\lambda_b$ are weighting terms for the clean prototype and binary outlier classification losses, respectively.

\vspace{-2mm}
\section{Experiments}
\label{sec:experiments}
\vspace{-1mm}
\subsection{Experimental setup}\label{sec:setup}
\vspace{-2mm}
\minisection{Datasets} We experiment on two common FSL datasets: MiniImageNet~\cite{vinyals2016matching} and TieredImageNet~\cite{ren2018meta}. Both include $84\times84$ pixel images. MiniImageNet contains 64, 16, and 20 classes for train, validation, and test, with 60K images in total. TieredImageNet consists of 351, 97, and 160 classes for train, validation, and test, with $\sim$0.78M images in total.

\minisection{Label noise types} We explore three forms of label noise: 
\vspace{-1mm}
\begin{tight_itemize}
    \item{\em Symmetric label swap} noise~\cite{van2015learning} draws mislabeled samples, uniformly at random, from the other $N-1$ classes of the episode, with the restriction that a noisy class does not tie or outnumber the original clean class.
    \item{\em Paired label swap} noise~\cite{han2018co} is more challenging: we always draw mislabeled samples from the same class by assigning each class with a {\em noisy class} counterpart, simulating real-world tendencies to confuse certain classes with others during labeling. We randomly generate these assignments in each episode as a derangement, to prevent models from learning these pairings across episodes.
    \item{\em Outlier} noise is sampled from classes {\em outside} the $N$-way episode. We use 600 images from each of the 350 ImageNet classes unincluded from MiniImageNet and TieredImageNet. We split these classes in half for meta-training and meta-test, to ensure that meta-test episode outliers represent previously unseen classes.
\end{tight_itemize}
\vspace{-1mm}

\begin{table}[t]
\centering
\caption{
{\bf Few-shot with paired label swap noise}. 5-way 5-shot Acc. $\pm$ 95\% CI on {\textcolor{blue}{MiniImageNet}}~\cite{vinyals2016matching}, {\textcolor{darkgreen}{TieredImageNet}}~\cite{ren2018meta}.
}
\vspace{-3mm}
\resizebox{0.85\columnwidth}{!}{
\begin{tabular}{c c || j k } 
\toprule[1.2pt]
& Model \textbackslash ~ Noise Proportion & \multicolumn{2}{c}{$40\%$} \\
\toprule[0.8pt]		
& Oracle             & 62.60 $\pm$ 0.17 & 66.14 $\pm$ 0.20  \\
\hline
{\multirow{9}{*}{\rotatebox[origin=c]{90}{Baselines}}} 
& Nearest $k=1$      & 37.97 $\pm$ 0.17 & 39.40 $\pm$ 0.18  \\
& Nearest $k=3$      & 37.84 $\pm$ 0.16 & 39.70 $\pm$ 0.18  \\
& Nearest $k=5$      & 40.39 $\pm$ 0.17 & 42.17 $\pm$ 0.18  \\
& Linear Classifier  & 44.49 $\pm$ 0.17 & 46.70 $\pm$ 0.18  \\
& Matching Networks~\cite{vinyals2016matching} & 43.53 $\pm$ 0.17 & 46.13 $\pm$ 0.19  \\
& MAML~\cite{finn2017model} & 40.67 $\pm$ 0.18 & 41.66 $\pm$ 0.18  \\
& Vanilla ProtoNet~\cite{snell2017prototypical} & 47.77 $\pm$ 0.19 & 50.85 $\pm$ 0.21  \\
& Baseline++~\cite{chen2019closer} & 47.82 $\pm$ 0.18 & 50.69 $\pm$ 0.20 \\
& RNNP~\cite{mazumder2021rnnp} & 47.88 $\pm$ 0.19 & 50.91 $\pm$ 0.20  \\
\hline
{\multirow{6}{*}{\rotatebox[origin=c]{90}{Ours}}}
& Median             & 48.81 $\pm$ 0.19 & 51.91 $\pm$ 0.21  \\
& Absolute   & 49.38 $\pm$ 0.20 & 52.40 $\pm$ 0.22 \\
& Euclidean  & 48.67 $\pm$ 0.19 & 51.90 $\pm$ 0.21 \\
&Cosine      & 49.40 $\pm$ 0.19 & 52.72 $\pm$ 0.22  \\
\cmidrule[0.8pt]{2-4}
& TraNFS-$2$	 & 50.63 $\pm$ 0.22 & 54.82 $\pm$ 0.24 \\
& TraNFS-$3$   & \textbf{53.96 $\pm$ 0.23} &\textbf{55.12 $\pm$ 0.24} \\
\bottomrule[1.2pt]
\end{tabular}
}
\vspace{-4mm}
\label{tab:pair_swap_5w5k}
\end{table}

\begin{table*}[t!]
\centering
\setlength{\tabcolsep}{6pt}
\def\arraystretch{1.05}
\caption{
{\bf Few-shot with outlier noise}. 5-way 5-shot Acc. $\pm$ 95\% CI on {\textcolor{blue}{MiniImageNet}}~\cite{vinyals2016matching}, {\textcolor{darkgreen}{TieredImageNet}}~\cite{ren2018meta}.
Our TraNSF is comparable to existing methods at $0\%$ or low noise, with a growing gap in its favor as noise levels increase.
Best viewed in color.
}
\vspace{-3mm}
\resizebox{0.95\textwidth}{!}{
\begin{tabular}{c c || j k j k j k j k} 
    \toprule[1.2pt]
& Model \textbackslash ~ Noise Proportion & \multicolumn{2}{c}{$0\%$} & \multicolumn{2}{c}{$20\%$} & \multicolumn{2}{c}{$40\%$} & \multicolumn{2}{c}{$60\%$} \\
\toprule[0.8pt]		
& Oracle             & 68.18 $\pm$ 0.16 & 71.42 $\pm$ 0.18 & 66.08 $\pm$ 0.17 & 69.19 $\pm$ 0.19 & 62.60 $\pm$ 0.17 & 66.14 $\pm$ 0.20 & 56.89 $\pm$ 0.18 & 60.39 $\pm$ 0.21 \\
\hline
{\multirow{9}{*}{\rotatebox[origin=c]{90}{Baselines}}} &
Nearest $k=1$      & 55.87 $\pm$ 0.18 & 58.89 $\pm$ 0.20 & 50.90 $\pm$ 0.18 & 54.57 $\pm$ 0.20 & 45.28 $\pm$ 0.18 & 49.45 $\pm$ 0.20 & 38.75 $\pm$ 0.18 & 43.20 $\pm$ 0.19 \\
& Nearest $k=3$      & 55.28 $\pm$ 0.18 & 58.38 $\pm$ 0.20 & 50.53 $\pm$ 0.17 & 53.98 $\pm$ 0.20 & 44.40 $\pm$ 0.17 & 48.06 $\pm$ 0.19 & 37.03 $\pm$ 0.16 & 40.11 $\pm$ 0.18 \\
& Nearest $k=5$      & 56.34 $\pm$ 0.17 & 59.25 $\pm$ 0.19 & 52.32 $\pm$ 0.17 & 55.30 $\pm$ 0.19 & 46.49 $\pm$ 0.17 & 49.34 $\pm$ 0.19 & 38.44 $\pm$ 0.16 & 40.56 $\pm$ 0.17 \\
& Linear Classifier  & 66.70 $\pm$ 0.16 & 69.60 $\pm$ 0.18 & 61.13 $\pm$ 0.17 & 64.58 $\pm$ 0.19 & 53.86 $\pm$ 0.18 & 57.57 $\pm$ 0.20 & 44.05 $\pm$ 0.18 & 47.90 $\pm$ 0.20 \\
& Matching Networks~\cite{vinyals2016matching} & 62.05 $\pm$ 0.17 & 64.99 $\pm$ 0.19 & 57.69 $\pm$ 0.18 & 60.74 $\pm$ 0.20 & 51.32 $\pm$ 0.19 & 54.28 $\pm$ 0.21 & 42.39 $\pm$ 0.19 & 44.93 $\pm$ 0.20 \\
& MAML~\cite{finn2017model} & 63.21 $\pm$ 0.18 & 63.90 $\pm$ 0.19 & 57.35 $\pm$ 0.19 & 58.14 $\pm$ 0.19 & 50.00 $\pm$ 0.19 & 51.11 $\pm$ 0.20 & 40.90 $\pm$ 0.17 & 42.01 $\pm$ 0.20 \\
& Vanilla ProtoNet~\cite{snell2017prototypical} & 68.18 $\pm$ 0.16 & \textbf{71.42 $\pm$ 0.18} & 63.92 $\pm$ 0.17 & 67.58 $\pm$ 0.19 & 57.07 $\pm$ 0.18 & 60.97 $\pm$ 0.20 & 46.99 $\pm$ 0.20 & 50.29 $\pm$ 0.21 \\
& Baseline++~\cite{chen2019closer} & 67.85 $\pm$ 0.16 & 71.29 $\pm$ 0.18 & 63.49 $\pm$ 0.17 & 67.07 $\pm$ 0.19 & 56.84 $\pm$ 0.18 & 60.64 $\pm$ 0.20 & 46.96 $\pm$ 0.19 & 50.07 $\pm$ 0.21 \\
& RNNP~\cite{mazumder2021rnnp} & 68.17 $\pm$ 0.16 & 71.28 $\pm$ 0.18 & 63.80 $\pm$ 0.17 & 67.29 $\pm$ 0.19 & 56.97 $\pm$ 0.18 & 60.83 $\pm$ 0.20 & 46.92 $\pm$ 0.20 & 50.09 $\pm$ 0.21 \\
\hline
{\multirow{6}{*}{\rotatebox[origin=c]{90}{Ours}}}
& Median     & 68.37 $\pm$ 0.16 & 71.28 $\pm$ 0.18 & 64.46 $\pm$ 0.17 & 67.79 $\pm$ 0.19 & 57.85 $\pm$ 0.18 & 61.63 $\pm$ 0.21 & 47.19 $\pm$ 0.20 & 50.63 $\pm$ 0.21 \\
& Absolute   & 68.13 $\pm$ 0.16 & 71.17 $\pm$ 0.18 & 64.69 $\pm$ 0.17 & \textbf{68.00 $\pm$ 0.19} & 58.30 $\pm$ 0.18 & 61.98 $\pm$ 0.21 & 47.39 $\pm$ 0.20 & 50.59 $\pm$ 0.22 \\
& Euclidean  & \textbf{68.51 $\pm$ 0.16} & 71.28 $\pm$ 0.18 & 64.57 $\pm$ 0.17 & 67.89 $\pm$ 0.19 & 58.01 $\pm$ 0.18 & 61.61 $\pm$ 0.20 & 47.25 $\pm$ 0.20 & 50.49 $\pm$ 0.21 \\
& Cosine      & 68.20 $\pm$ 0.16 & 70.79 $\pm$ 0.18 & 64.78 $\pm$ 0.17 & 67.94 $\pm$ 0.19 & 58.36 $\pm$ 0.18 & 62.37 $\pm$ 0.21 & 47.34 $\pm$ 0.20 & 51.12 $\pm$ 0.22 \\
\cmidrule[0.8pt]{2-10}
& TraNFS-$2$	     & 67.76 $\pm$ 0.17 & 70.83 $\pm$ 0.19 & 64.47 $\pm$ 0.19 & 67.52 $\pm$ 0.21 & 58.29 $\pm$ 0.20 & 61.76 $\pm$ 0.22 & 47.37 $\pm$ 0.23 & 51.40 $\pm$ 0.23 \\
& TraNFS-$3$         & 68.11 $\pm$ 0.17 & 71.13 $\pm$ 0.19 & \textbf{64.96 $\pm$ 0.18} & 67.93 $\pm$ 0.20 & \textbf{59.03 $\pm$ 0.20} & \textbf{62.39 $\pm$ 0.22} & \textbf{47.69 $\pm$ 0.22} & \textbf{51.82 $\pm$ 0.23} \\
\bottomrule[1.2pt]
\end{tabular}
}
\vspace{-4mm}
\label{tab:outlier_5w5k}
\end{table*}

The amount of noise in a support set is specified as the percent of the total number of shots. We only consider settings where the clean class can reasonably be identified. Thus, for example, we only consider paired label swap noise under $50\%$, as at $50\%$ noise and above, the clean class is ambiguous or a minority. We also exclude paired label swap settings that are identical to the corresponding symmetric label swap setting (\eg $20\%$ for 5-shot 5-way).

\minisection{Model} Our models are implemented in PyTorch~\cite{paszke2019pytorch}, using learn2learn~\cite{arnold2020learn2learn} as a starting point.
We set $T=25$ for similarity weighted prototypes with squared euclidean and absolute distances and $T=0.2$ for cosine similarity.
For TraNFS, we instantiate the transformer $\mathcal T$ using 2 or 3-layers with eight heads, learnable positional embeddings $\texttt{POS}^{(c)}$, and random constant class tokens $\texttt{CLS}^{(c)}$. 
We apply an orthogonally initialized pair of down-projection and up-projection weight matrices before and after the transformer, reducing the transformer's dimensionality to 128. 
We found these orthogonal projections stabilize training~\cite{rangrej2021revisiting}, while also greatly reducing the number of transformer parameters. 
Finally, we set the hyperparameters of Eq.~\eqref{eq:loss} as $\lambda_b = 0.5$ and $\lambda_c = 5$. 
See Appx.~\ref{apx:trans_abl} for hyperparameter sweeps.
Note that while the transformer must be meta-trained and used to generate robust prototypes during meta-test, it is not used during inference on individual query samples. Hence, the number of parameters and computation cost during inference is similar to methods like ProtoNet~\cite{snell2017prototypical}.

\minisection{Training and testing} To isolate the effect of the method from the learned features, we use the same, frozen, 4-layer convolutional backbone~\cite{vinyals2016matching}, trained with the ProtoNet objective, for all models except MAML~\cite{finn2017model}. We chose this simple backbone to emphasize the method, not the feature extractor. The backbone is trained with AdamW~\cite{loshchilov2018decoupled}, with weight decay of $0.01$, initial learning rate $1\times 10^{-3}$, and learning rate decay of $\times 0.7$ every 10K episodes for 100K episodes for MiniImageNet, and every 25K episodes for 250K episodes for TieredImageNet. Meta-validation is used for accuracy model selection. Our TraNFS is similarly optimized, with initial learning rate $5\times 10^{-4}$, decayed after every 25K episodes, for 200K episodes. We use random horizontal flips, resized crops, and color jitters as data augmentations for all models. Finally, each meta-train and meta-test episode has 15 queries. We report mean accuracy and 95\% confidence interval for 10K meta-test episodes. All experiments are run on a single Nvidia V100 GPU. 

\begin{figure*}[t]
  \centering
  \hfill
  \begin{subfigure}{0.3\linewidth}
    \centering
    \includegraphics[width=0.81\textwidth]{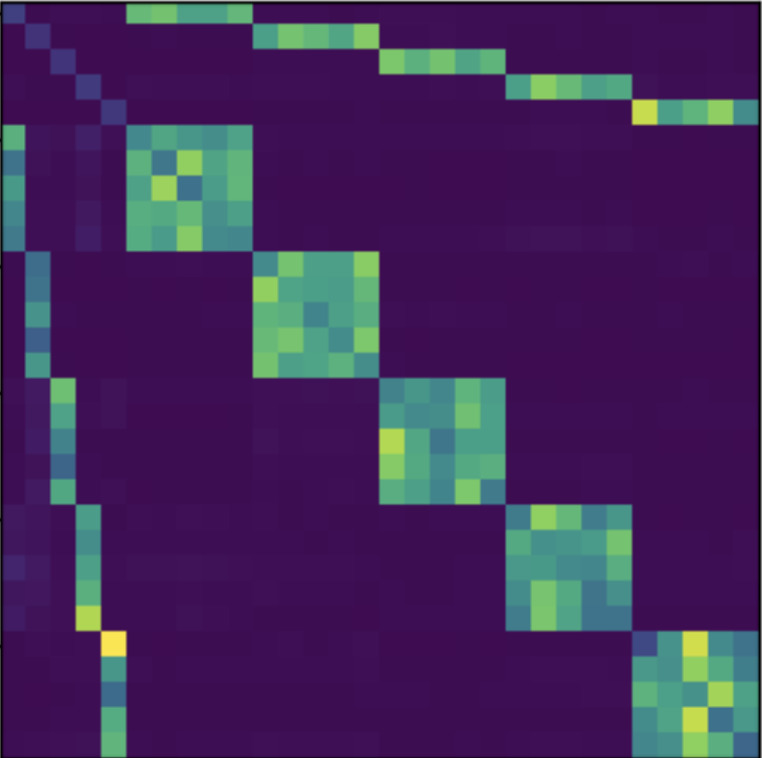}
    \caption{Layer 1}
    \label{fig:attn_map1}
  \end{subfigure}
  \hfill
  \begin{subfigure}{0.3\linewidth}
    \centering
    \includegraphics[width=0.81\textwidth]{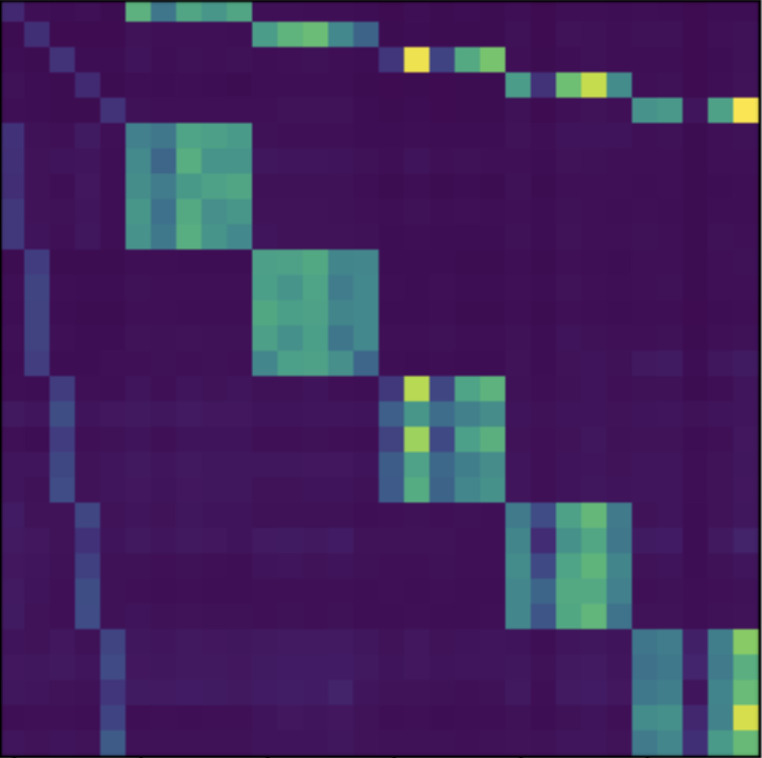}
    \caption{Layer 2}
    \label{fig:attn_map2}
  \end{subfigure}
  \hfill
  \begin{subfigure}{0.3\linewidth}
    \centering
    \includegraphics[width=0.81\textwidth]{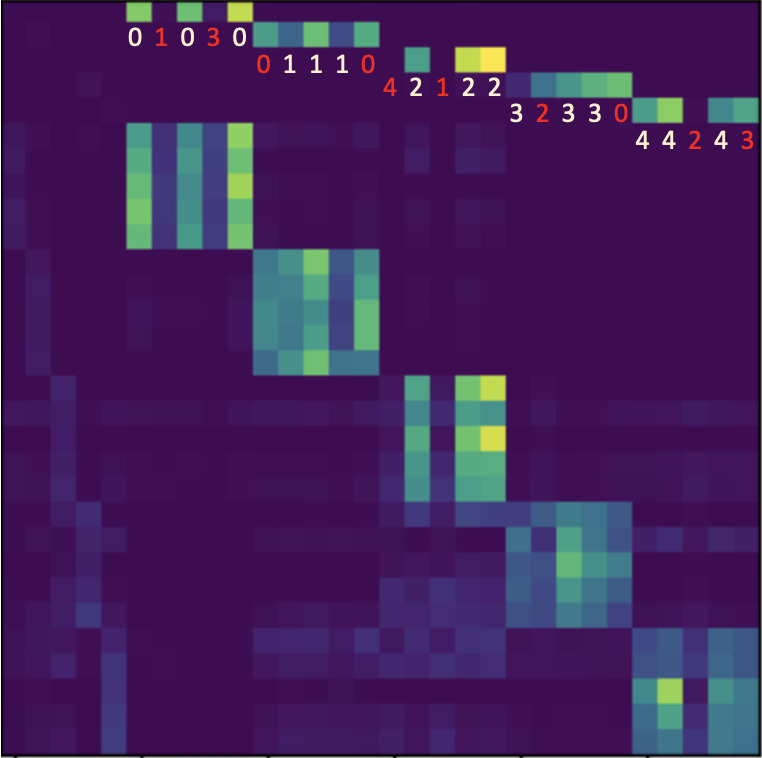}
    \caption{Layer 3}
    \label{fig:attn_map3}
  \end{subfigure}
  \hfill
\vspace{-3mm}
\caption{{\bf Attention maps from selected TraNFS-3 self-attention heads.} 5-way 5-shot MiniImageNet, $40\%$ symmetric noise. The first five rows of each map correspond to positions of \texttt{CLS} tokens, which produce the output prototypes. True class label of each support sample overlaid on the final layer's attention maps. Evidently, later layers assign lower attention to noisy samples, effectively filtering~them.
}
\vspace{-4mm}
\label{fig:attn_maps}
\end{figure*}

\subsection{Noisy few-shot results}
\label{sec:nfs_res}
\vspace{-2mm}
We compare all our proposed methods for noisy few-shot learning--Median, Absolute, Euclidean, Cosine, and TraNFS--with several baselines (see Appx.~\ref{apx:baselines} for details on these baselines). We report results for 5-way and 5-shot\footnote{See Appx.~\ref{apx:5w_K} for 3-shot and 10-shot experiments.} on MiniImageNet and TieredImageNet using symmetric noise (Table~\ref{tab:sym_swap_5w5k}), paired noise (Table~\ref{tab:pair_swap_5w5k}), and outlier noise (Table~\ref{tab:outlier_5w5k}). 
We further report results for an {\em Oracle}: a ProtoNet~\cite{snell2017prototypical} that knows which samples are mislabeled and ignores them by removing them from each support set, thereby representing perfect noise rejection (the blue line in Fig.~\ref{fig:proto_noisy}). 

Unsurprisingly, noisy labels negatively affect all methods. Our proposed median and similarity weighting alternatives to ProtoNet's mean suffer less than the baselines, on all three noise types. This is expected, as their aggregation methods are less sensitive to outliers. Furthermore, our transformer-based TraNFS is clearly superior to its baselines. For example, consider the challenging 5-way 5-shot setting on MiniImageNet with $40\%$ paired label swap noise. Our TraNFS provides a $6.19\%$ absolute improvement in accuracy over ProtoNet, representing {\em a significant relative drop of $41.7\%$ in error}, compared with the Oracle. As explained in Sec.~\ref{sec:attn_maps}, this gain is due to the transformer's self-attention learning to compare support set examples and suppress samples suspected of being mislabeled. 

Comparing noise types, we find that, as expected, FSL methods are more vulnerable to paired label swap than symmetric noise.
Outlier noise has the least impact on model performance, with three outlier samples in a 5-way 5-shot test reducing accuracy similarly to two label swap samples. We reason that this is due to the direction in which these noisy samples push the model's decision boundary: label swapped samples pull the decision boundary closer to the features of other classes in the $N$-way classification task; for paired label swaps, this effort is coordinated across noisy samples, amplifying the effect.
In contrast, outlier samples have lower probability of being arranged in regions of feature space that interfere with the $N$-way classification.

\begin{table}[t!]
\centering
\caption{
{\bf Various amounts of injected meta-training artificial noise.} TraNFS-3 5-way 5-shot Acc. $\pm$ 95\% CI on MiniImageNet.
}
\vspace{-3mm}
\resizebox{\columnwidth}{!}{
\begin{tabular}{c c c c || c c c c} 
\toprule[1.2pt]
$0\%$ & $20\%$ & $40\%$ & $60\%$ & $0\%$ & $20\%$ & $40\%$ & $60\%$ \\
\toprule[0.8pt]		
\checkmark & & &                                   & \textbf{69.10 $\pm$ 0.16} & 63.56 $\pm$ 0.18 & 52.85 $\pm$ 0.19 & 39.19 $\pm$ 0.21 \\
& \checkmark & &                                   & 68.67 $\pm$ 0.17 & 64.85 $\pm$ 0.18 & 55.76 $\pm$ 0.21 & 41.73 $\pm$ 0.23 \\
& & \checkmark &                                   & 67.37 $\pm$ 0.17 & 63.97 $\pm$ 0.19 & 55.65 $\pm$ 0.21 & 41.63 $\pm$ 0.24 \\
& & & \checkmark                                   & 50.40 $\pm$ 0.19 & 48.26 $\pm$ 0.19 & 43.11 $\pm$ 0.21 & 35.44 $\pm$ 0.23 \\
& \checkmark & \checkmark &                        & 68.53 $\pm$ 0.17 & \textbf{65.08 $\pm$ 0.18} & 56.65 $\pm$ 0.21 & 42.60 $\pm$ 0.24 \\
\checkmark & \checkmark & \checkmark &             & 68.90 $\pm$ 0.17 & \textbf{65.08 $\pm$ 0.18} & \textbf{56.73 $\pm$ 0.21} & \textbf{42.69 $\pm$ 0.24} \\
& \checkmark & \checkmark & \checkmark             & 66.92 $\pm$ 0.17 & 63.52 $\pm$ 0.19 & 54.98 $\pm$ 0.22 & 42.01 $\pm$ 0.24 \\
\checkmark & \checkmark & \checkmark & \checkmark  & 67.64 $\pm$ 0.17 & 63.83 $\pm$ 0.18 & 54.81 $\pm$ 0.21 & 41.33 $\pm$ 0.24 \\
\bottomrule[1.2pt]
\end{tabular}
}
\vspace{-4mm}
\label{tab:noise_prop}
\end{table}

\begin{table}[t!]
\centering
\caption{
{\bf Meta-training mean/median prototype models with noise.} 5-way 5-shot Acc. $\pm$ 95\% CI on MiniImageNet. 
}
\vspace{-3mm}
\resizebox{\columnwidth}{!}{
\begin{tabular}{c || c c c c} 
\toprule[1.2pt]
Baseline ~ \textbackslash ~ Noise Proportion & $0\%$ & $20\%$ & $40\%$ & $60\%$ \\
\toprule[0.8pt]		
Mean + Sym (0\%, 20\%, 40\%)    & 67.89 $\pm$ 0.16 & 62.44 $\pm$ 0.18 & 51.66 $\pm$ 0.19 & 38.53 $\pm$ 0.20 \\
Mean + Pair (0\%, 20\%, 40\%)   & -            & -            & 48.05 $\pm$ 0.19 & -            \\
Mean + Out (0\%, 20\%, 40\%)    & 66.88 $\pm$ 0.16 & 62.69 $\pm$ 0.17 & 56.00 $\pm$ 0.18 & 45.90 $\pm$ 0.19 \\
Median + Sym (0\%, 20\%, 40\%)  & 67.11 $\pm$ 0.16 & 62.39 $\pm$ 0.17 & 51.70 $\pm$ 0.19 & 39.57 $\pm$ 0.20 \\
Median + Pair (0\%, 20\%, 40\%) & -            & -            & 48.64 $\pm$ 0.19 & -            \\
Median + Out (0\%, 20\%, 40\%)  & 67.17 $\pm$ 0.16 & 63.46 $\pm$ 0.17 & 57.01 $\pm$ 0.18 & 46.44 $\pm$ 0.20 \\
\bottomrule[1.2pt]
\end{tabular}
}
\vspace{-4mm}
\label{tab:baseline_noise}
\end{table}

\subsection{Visualizing transformer attention to noise}
\label{sec:attn_maps}
\vspace{-2mm}
To understand how our proposed transformer model suppresses noisy samples, we visualize the self-attention from selected attention heads at each of its layers. The attention maps for a 5-way 5-shot test episode of MiniImageNet with 40\% symmetric label swap noise are shown in Fig.~\ref{fig:attn_maps}. From Fig.~\ref{fig:attn_map1}, attention at the \texttt{CLS} token positions suggests that the first layer of the transformer mainly uses positional encodings to focus on per-class examples, resulting in a representation reminiscent of ProtoNet's mean. 

Subsequent layers are visualized in  Fig.~\ref{fig:attn_map2}-\ref{fig:attn_map3}. Evidently, the transformer is able to refine its class representations by decreasing attention to noisy samples, suppressing their influence on the aggregated representations. For example, our self-attention mechanism correctly learned to ignore the $2^{\mathrm{nd}}$ and $4^{\mathrm{th}}$ samples of the first class, which are indeed mislabeled. While this filtering ability is not perfect, we emphasize that learning class concepts from so few samples, without any prior concept of the class, is a challenging task (see Fig.~\ref{fig:qualitative}); ImageNet contains enough intra-class variation and label ambiguity that identifying mislabeled samples can be challenging even for humans, who have the advantage of conceptual priors of the ImageNet classes.

\subsection{Ablations: Meta-training noise proportion}\label{sec:noise_prop}
\vspace{-2mm}
To test the influence of adding training noise,\footnote{See Appx.~\ref{apx:trans_abl} for more ablation studies on other design choices.} we meta-train a 3-layer TraNFS model on 5-shot, 5-way MiniImageNet with various amounts of symmetric label swap noise, synthetically added to meta-training. Table~\ref{tab:noise_prop} reports these results, clearly showing a few patterns.

First, training with a single noise percentage boosts performance on that noise level during meta-test. 
Those models, however, do not generalize well to other noise levels. 
Instead, training on varying noise levels seems to offer the best results across a range of meta-test support set noise levels. 
This is important, as the stochastic nature of label noise means we expect real world support sets to have varying levels of noise, and it is desirable to handle multiple noise levels with a single model.
In particular, training on support sets with $\{0, 20, 40\}\%$ appears to achieve the best overall performance. 
Finally, we observe that training on extremely noisy support sets (\eg, $60\%$) appears counter-productive. 
We believe that this is due to a mixture of having a more challenging task to learn while also diluting learnable information of the clean class of each support set.

Synthetically adding noise during meta-training proved essential for TraNFS. A similar strategy could conceivably also be applied other methods. As Table~\ref{tab:baseline_noise} shows, however, this approach was unhelpful. We believe that the absence of learnable mechanisms for rejecting noise encourages these baselines to learn strong feature extractors on the highest quality (\ie, cleanest) data available. Thus, we do not add artificial noise to our baselines during meta-training.

\section{Conclusion}
\label{sec:conclusion}
\vspace{-2mm}
We focus on a key vulnerability of modern FSL methods: noisy, mislabeled support sets samples. We propose several technical novelties to mitigate this vulnerability: replacing the mean aggregator used by ProtoNets with a median or similarity weighted aggregation. We then present a novel, transformer-based model designed to learn a dynamic noise rejection mechanism, leveraging the transformer's attention mechanism. Experiments on MiniImageNet and TieredImageNet under varying types and levels of noise clearly show the effectiveness of our techniques.

\minisection{Limitations} As with other FSL methods, we assume a cleanly labeled meta-train dataset with noisy labels mostly affecting meta-testing. We believe this is reasonable: collecting meta-train data and meta-training are performed offline, before model deployment, with reasonable control over both.
Sometimes, however, meta-training datasets can also be noisy. While we introduce noise during meta-training, our methods assume that the query set is correctly labeled. Queries with noisy labels could cause misleading gradients. In such cases, ideas from the noisy label literature~\cite{jiang2018mentornet, han2018co} could offer promising direction for future work.

{\small
\bibliographystyle{ieee_fullname}
\bibliography{bib}
}

\clearpage
\appendix

\begin{figure*}[t]
  \centering
  \hfill
  \begin{subfigure}{0.32\linewidth}
    \centering
    \includegraphics[width=\textwidth]{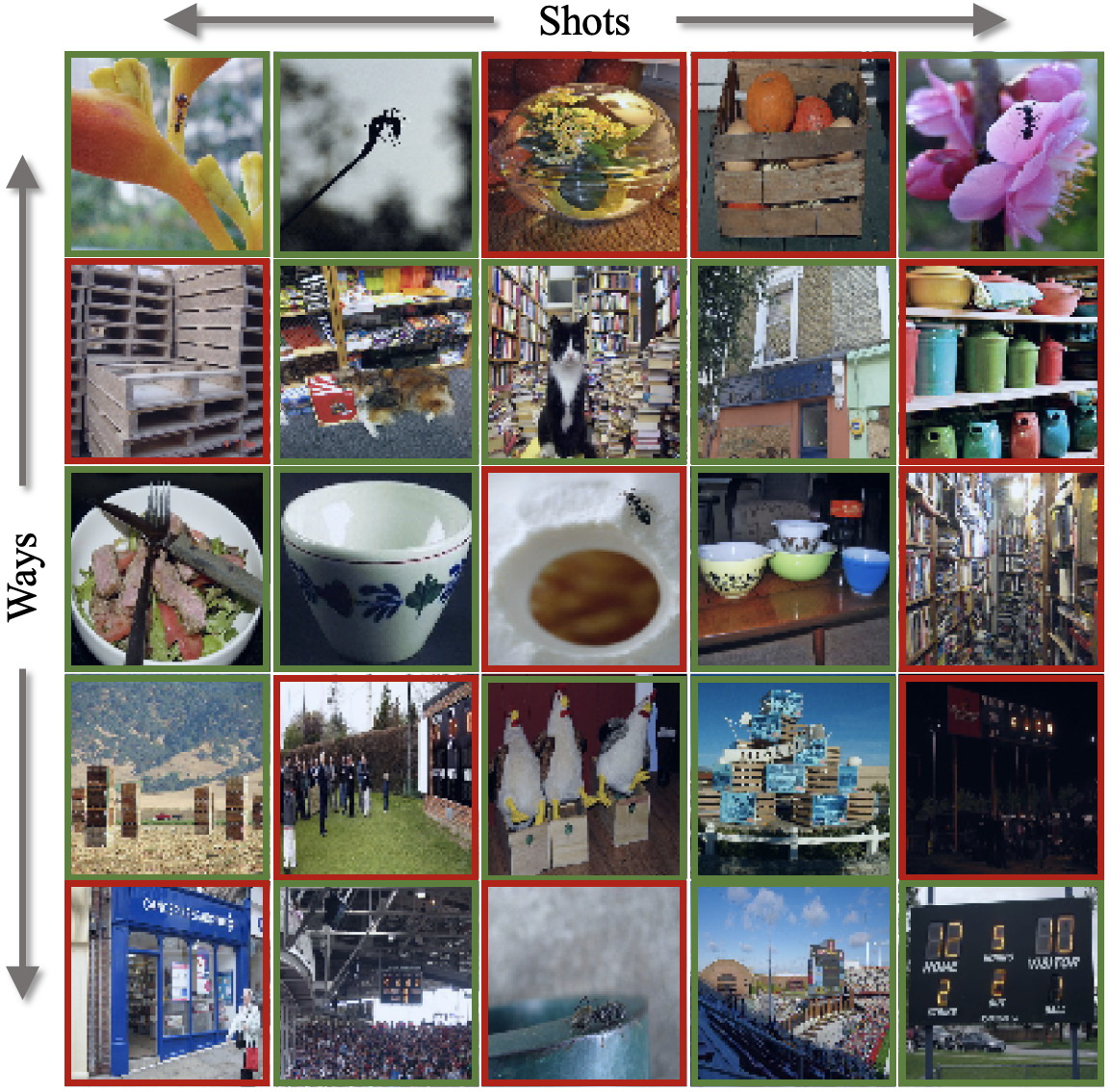}
    \caption{}
  \end{subfigure}
  \hfill
  \begin{subfigure}{0.32\linewidth}
    \centering
    \includegraphics[width=\textwidth]{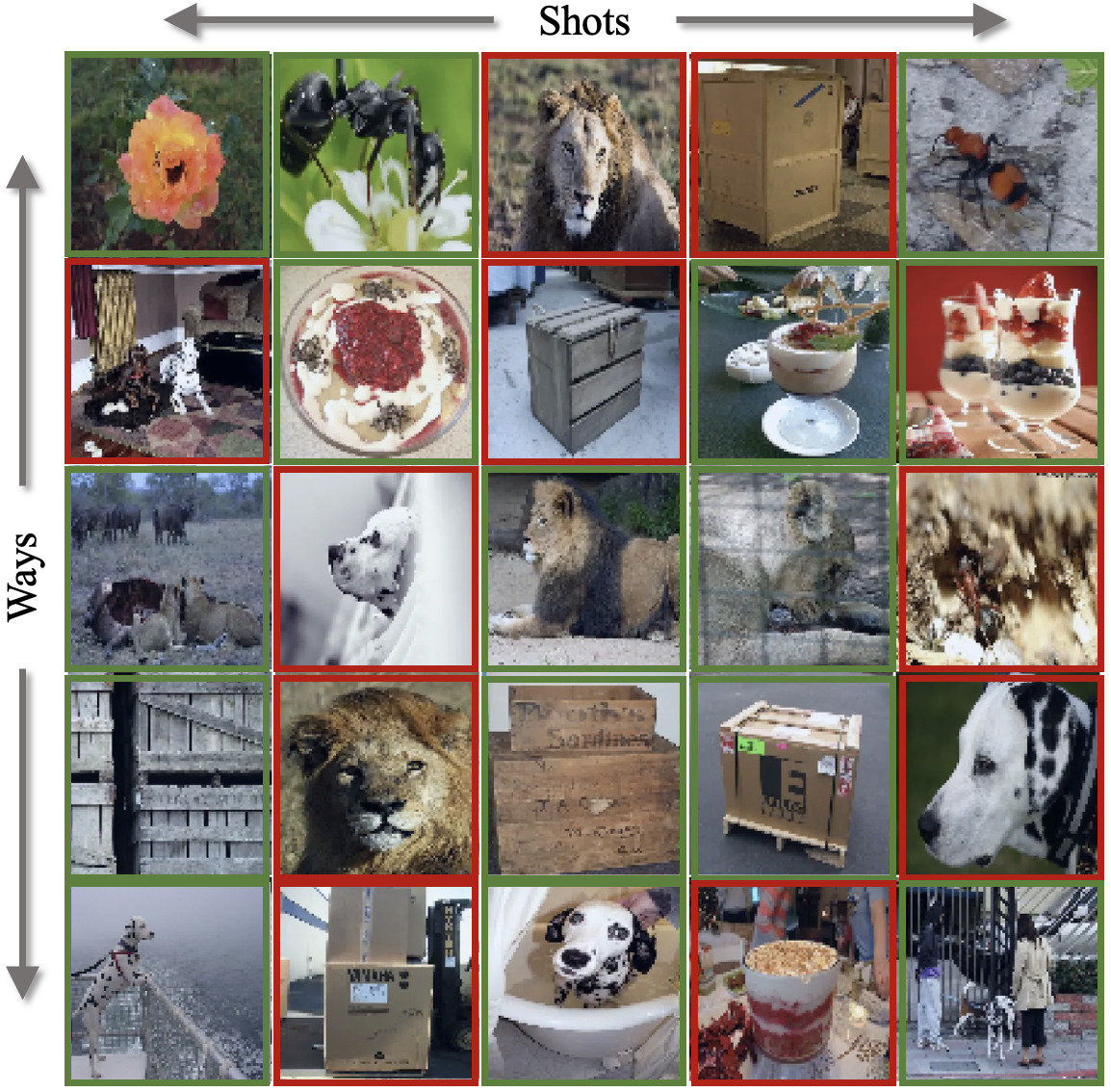}
    \caption{}
  \end{subfigure}
  \hfill
  \begin{subfigure}{0.32\linewidth}
    \centering
    \includegraphics[width=\textwidth]{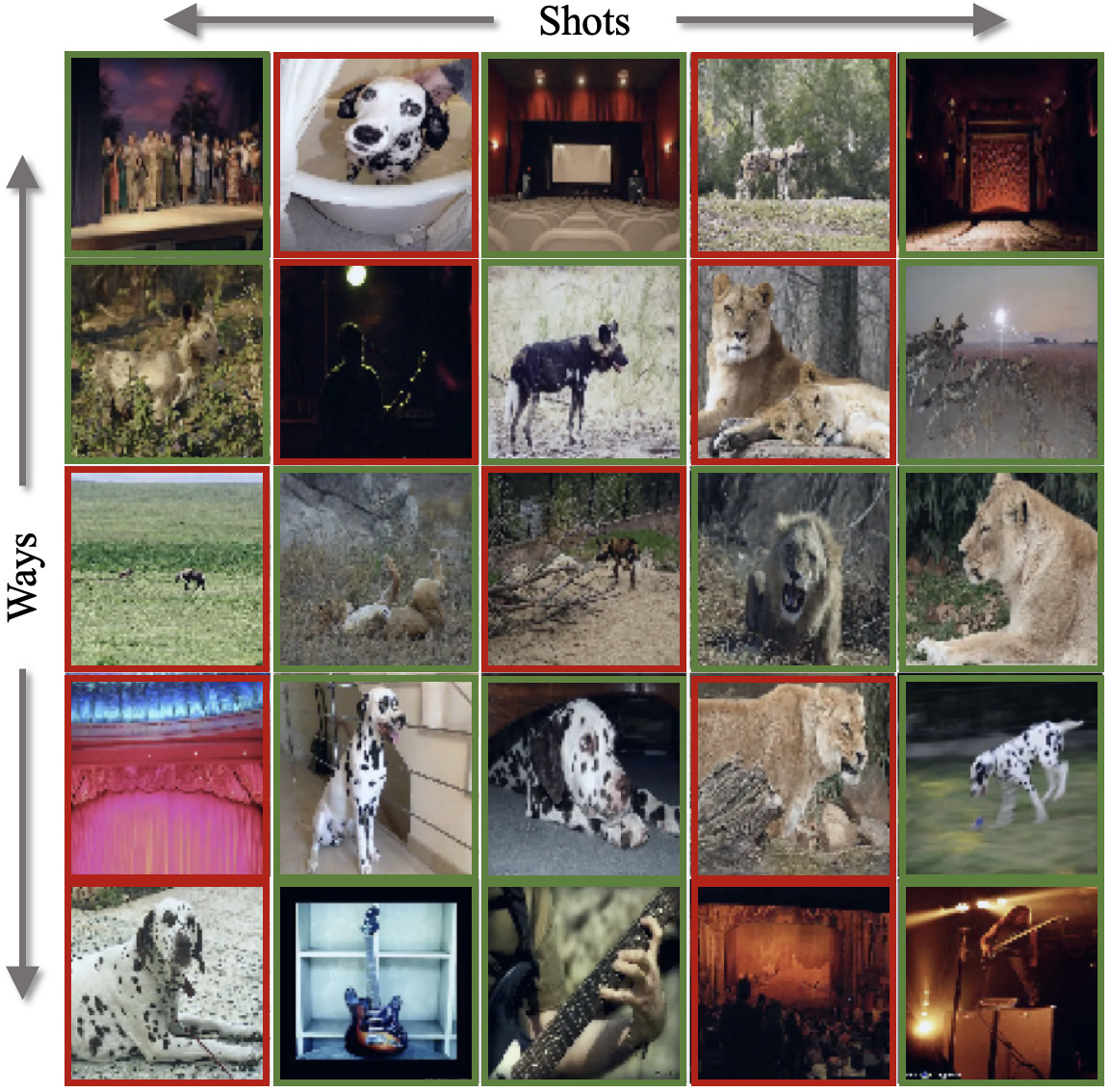}
    \caption{}
  \end{subfigure}
  \hfill
\vspace{-3mm}
\caption{{\bf Noisy support set examples.} Images with green boxes are clean samples from the original class, while red boxes are mislabeled samples due to symmetric label swaps. (a) is the support set shown in Fig.~\ref{fig:qualitative} of the main paper.}
\label{fig:examples}
\end{figure*}

We include supplemental material for our work here.
Appx.~\ref{apx:nfsl_ex} shows the mislabeled samples in the 5-way 5-shot support set in Fig.~\ref{fig:qualitative}, as well as two more example noisy support sets.
We discuss computational complexity considerations for iteratively solving for the median in Sec.~\ref{apx:median}.
In Appx.~\ref{apx:5w_K}, we investigate noisy few-shot performance for different numbers of shots from the 5-shot setting considered in Sec.~\ref{sec:nfs_res}.
Appx.~\ref{apx:baselines} contains descriptions and additional implementation details of the baselines that we compare against.
We perform further ablation studies beyond Sec.~\ref{sec:noise_prop}, investigating feature extractors, hyperparameter settings, and architectural design choices of TraNFS in Appx.~\ref{apx:trans_abl}.

\section{Noisy support set examples}
\label{apx:nfsl_ex}
Noisy few-shot learning is a challenging problem.
Even before adding noise, there can be significant variation within a class, largely due to the manner in which the ImageNet~\cite{deng2009imagenet} dataset (from which MiniImageNet~\cite{vinyals2016matching} and TieredImageNet~\cite{ren2018meta} are derived) was constructed.
Some images in the \textit{clean} version of ImageNet are mislabeled due to human error, but even among the correctly labeled objects, there are non-canonical views, images with multiple objects (possibly from multiple ImageNet classes), and classes that are close to synonymous.
We provide several examples of noisy support sets from MiniImageNet with $40\%$ symmetric label swap noise~\cite{van2015learning} in Fig.~\ref{fig:examples}, with the clean and noisy samples framed in green and red, respectively.
While humans are generally able to separate the noisy samples from the clean samples with some scrutiny, this is in large part due to prior conceptual understandings of the classes depicted.
Few-shot models presented with support sets such as those in Fig.~\ref{fig:examples} are tasked with learning how to distinguish the depicted classes \textit{without having previously seen these classes}, a much more difficult problem.

\section{A Note On Median Complexity}
\label{apx:median}
As discussed in Sec.~\ref{sec:median}, median computation has to be performed iteratively since no closed form solution exists. We have chosen the $2^{\mathrm{nd}}$- over $1^{\mathrm{st}}$-order optimization as the former provides an optimal step size at
each iteration, speeding up convergence. This choice may seem costly at first glance, but computational complexity analysis of Eq.~\eqref{eq:medupdate} shows negligible $2^{\mathrm{nd}}$-order method overhead. Each median update iteration takes $4DK + 2K - D$ flops for gradient computation, $K$ flops for optimal step calculation ($2^{\mathrm{nd}}$-order method overhead), and $2D$ flops for parameter update. We emphasize that this optimization is done to calculate a median prototype (as opposed to updating the model weights); $D$ and $K$ are both fairly small.

\begin{table}[t!]
\centering
\caption{
Few-shot performance with symmetric label swap noise on 5-way 3-shot MiniImageNet~\cite{vinyals2016matching}.
}
\setlength{\tabcolsep}{6pt}
\def\arraystretch{1.05}
\resizebox{0.8\columnwidth}{!}{
\begin{tabular}{c || c c} 
    \toprule[1.2pt]
Model \textbackslash ~ Noise Proportion & $0\%$ & $33.3\%$ \\
\toprule[0.8pt]		
Oracle             & 62.60 $\pm$ 0.17 & 56.89 $\pm$ 0.18 \\
\hline
Nearest $k=1$      & 52.98 $\pm$ 0.18 & 39.92 $\pm$ 0.18 \\
Nearest $k=3$      & 50.59 $\pm$ 0.18 & 38.76 $\pm$ 0.16 \\
Nearest $k=5$      & 50.20 $\pm$ 0.17 & 40.05 $\pm$ 0.16 \\
Linear Classifier  & 61.54 $\pm$ 0.17 & 46.06 $\pm$ 0.17 \\
Matching Networks~\cite{vinyals2016matching}  & 57.86 $\pm$ 0.18 & 44.92 $\pm$ 0.18 \\
MAML~\cite{finn2017model} & 59.79 $\pm$ 0.20 & 40.41 $\pm$ 0.17 \\
Vanilla ProtoNet~\cite{snell2017prototypical} & 62.54 $\pm$ 0.18 & 48.78 $\pm$ 0.19 \\
RNNP~\cite{mazumder2021rnnp} & 62.57 $\pm$ 0.17 & 48.76 $\pm$ 0.19 \\
\hline
Median             & 62.60 $\pm$ 0.17 & 50.40 $\pm$ 0.19 \\
Absolute $T=10.0$  & 61.77 $\pm$ 0.17 & 50.93 $\pm$ 0.19 \\
Absolute $T=25.0$  & 62.54 $\pm$ 0.17 & 50.84 $\pm$ 0.19 \\
Absolute $T=50.0$  & 62.69 $\pm$ 0.17 & 50.06 $\pm$ 0.19 \\
Euclidean $T=10.0$ & 62.58 $\pm$ 0.17 & 50.83 $\pm$ 0.19 \\
Euclidean $T=25.0$ & 62.62 $\pm$ 0.18 & 50.06 $\pm$ 0.19 \\
Euclidean $T=50.0$ & 62.62 $\pm$ 0.17 & 49.51 $\pm$ 0.19 \\
Cosine $T=0.2$     & 62.75 $\pm$ 0.17 & 49.63 $\pm$ 0.19 \\
Cosine $T=0.5$     & 62.55 $\pm$ 0.17 & 49.15 $\pm$ 0.19 \\
Cosine $T=1.0$     & 62.52 $\pm$ 0.17 & 49.20 $\pm$ 0.19 \\
Cosine $T=2.0$     & 62.63 $\pm$ 0.17 & 49.05 $\pm$ 0.19 \\
Cosine $T=5.0$     & 62.54 $\pm$ 0.17 & 48.96 $\pm$ 0.19 \\
\toprule[0.8pt]		
TraNFS-$2$	       & 64.17 $\pm$ 0.18 & 53.35 $\pm$ 0.21 \\
TraNFS-$3$         & \textbf{64.28 $\pm$ 0.18} & \textbf{53.84 $\pm$ 0.21} \\
\bottomrule[1.2pt]
\end{tabular}
}
\vspace{-2mm}
\label{tab:sym_swap_5w3k}
\end{table}

\begin{table}[t!]
\centering
\caption{
Few-shot performance with outlier noise on 5-way 3-shot MiniImageNet~\cite{vinyals2016matching}.
}
\setlength{\tabcolsep}{6pt}
\def\arraystretch{1.05}
\resizebox{0.8\columnwidth}{!}{
\begin{tabular}{c || c c} 
    \toprule[1.2pt]
Model \textbackslash ~ Noise Proportion & $0\%$ & $33.3\%$ \\
\toprule[0.8pt]		
Oracle             & 62.60 $\pm$ 0.17 & 56.89 $\pm$ 0.18 \\
\hline
Nearest $k=1$      & 53.07 $\pm$ 0.18 & 44.66 $\pm$ 0.18 \\
Nearest $k=3$      & 50.40 $\pm$ 0.18 & 41.59 $\pm$ 0.17 \\
Nearest $k=5$      & 50.24 $\pm$ 0.17 & 42.26 $\pm$ 0.17 \\
Linear Classifier  & 61.58 $\pm$ 0.17 & 51.21 $\pm$ 0.18 \\
Matching Networks~\cite{vinyals2016matching}  & 57.82 $\pm$ 0.18 & 48.56 $\pm$ 0.19 \\
MAML~\cite{finn2017model} & 59.76 $\pm$ 0.19 & 47.08 $\pm$ 0.19 \\
Vanilla ProtoNet~\cite{snell2017prototypical} & 62.43 $\pm$ 0.17 & 52.78 $\pm$ 0.19 \\
RNNP~\cite{mazumder2021rnnp} & 62.55 $\pm$ 0.17 & 52.88 $\pm$ 0.19 \\
\hline
Median             & 62.53 $\pm$ 0.17 & 53.82 $\pm$ 0.19 \\
Absolute $T=10.0$  & 61.54 $\pm$ 0.17 & 53.76 $\pm$ 0.19 \\
Absolute $T=25.0$  & 62.47 $\pm$ 0.17 & 54.07 $\pm$ 0.19 \\
Absolute $T=50.0$  & 62.69 $\pm$ 0.17 & 53.73 $\pm$ 0.19 \\
Euclidean $T=10.0$ & 62.56 $\pm$ 0.18 & 54.10 $\pm$ 0.19 \\
Euclidean $T=25.0$ & 62.57 $\pm$ 0.17 & 53.72 $\pm$ 0.18 \\
Euclidean $T=50.0$ & 62.76 $\pm$ 0.17 & 53.55 $\pm$ 0.19 \\
Cosine $T=0.2$     & 62.58 $\pm$ 0.17 & 53.46 $\pm$ 0.19 \\
Cosine $T=0.5$     & 62.50 $\pm$ 0.17 & 53.03 $\pm$ 0.19 \\
Cosine $T=1.0$     & 62.50 $\pm$ 0.17 & 52.84 $\pm$ 0.19 \\
Cosine $T=2.0$     & 62.72 $\pm$ 0.17 & 53.16 $\pm$ 0.19 \\
Cosine $T=5.0$     & 62.63 $\pm$ 0.18 & 53.19 $\pm$ 0.19 \\
\toprule[0.8pt]		
TraNFS-$2$	       & \textbf{63.63 $\pm$ 0.18} & \textbf{54.75 $\pm$ 0.20} \\
TraNFS-$3$         & 63.61 $\pm$ 0.18 & 54.72 $\pm$ 0.20 \\
\bottomrule[1.2pt]
\end{tabular}
}
\vspace{-2mm}
\label{tab:outlier_5w3k}
\end{table}

\section{Different number of shots}
\label{apx:5w_K}
While the experiments in Sec.~\ref{sec:nfs_res} are conducted with 5 shots, many of our findings on noisy FSL apply to other numbers of shots as well.
We provide additional results below for MiniImageNet~\cite{vinyals2016matching} with $K=\{3,10\}$ shots.

\begin{table*}[t!]
\centering
\caption{
Few-shot performance with symmetric label swap noise on 5-way 10-shot MiniImageNet~\cite{vinyals2016matching}.
}
\setlength{\tabcolsep}{6pt}
\def\arraystretch{1.05}
\resizebox{\textwidth}{!}{
\begin{tabular}{c || c c c c c c c c} 
    \toprule[1.2pt]
Model \textbackslash ~ Noise Proportion & $0\%$ & $10\%$ & $20\%$ & $30\%$ & $40\%$ & $50\%$ & $60\%$ & $70\%$ \\
\toprule[0.8pt]		
Oracle             & 73.62 $\pm$ 0.14 & 72.78 $\pm$ 0.15 & 71.78 $\pm$ 0.15 & 70.82 $\pm$ 0.15 & 69.27 $\pm$ 0.16 & 64.70 $\pm$ 0.17 & 60.59 $\pm$ 0.17 & 53.88 $\pm$ 0.18 \\
\hline
Nearest $k=1$      & 53.02 $\pm$ 0.19 & 49.04 $\pm$ 0.18 & 45.02 $\pm$ 0.18 & 40.87 $\pm$ 0.18 & 37.28 $\pm$ 0.17 & 33.13 $\pm$ 0.17 & 29.07 $\pm$ 0.16 & 24.64 $\pm$ 0.15 \\
Nearest $k=3$      & 53.79 $\pm$ 0.19 & 50.84 $\pm$ 0.18 & 47.24 $\pm$ 0.18 & 43.21 $\pm$ 0.17 & 38.58 $\pm$ 0.17 & 33.72 $\pm$ 0.16 & 29.00 $\pm$ 0.15 & 24.24 $\pm$ 0.13 \\
Nearest $k=5$      & 55.03 $\pm$ 0.20 & 53.08 $\pm$ 0.19 & 50.29 $\pm$ 0.19 & 46.50 $\pm$ 0.18 & 41.97 $\pm$ 0.17 & 36.51 $\pm$ 0.16 & 30.75 $\pm$ 0.15 & 25.26 $\pm$ 0.14 \\
Linear Classifier  & 72.08 $\pm$ 0.14 & 68.56 $\pm$ 0.15 & 64.17 $\pm$ 0.16 & 58.90 $\pm$ 0.16 & 52.68 $\pm$ 0.16 & 45.43 $\pm$ 0.16 & 37.20 $\pm$ 0.15 & 29.18 $\pm$ 0.14 \\
Matching Networks~\cite{vinyals2016matching} & 62.63 $\pm$ 0.19 & 60.81 $\pm$ 0.19 & 58.21 $\pm$ 0.19 & 54.79 $\pm$ 0.19 & 50.05 $\pm$ 0.19 & 43.47 $\pm$ 0.18 & 35.90 $\pm$ 0.17 & 28.70 $\pm$ 0.15 \\
MAML~\cite{finn2017model} & 64.37 $\pm$ 0.18 & 64.42 $\pm$ 0.18 & 55.27 $\pm$ 0.18 & 44.17 $\pm$ 0.18 & 44.10 $\pm$ 0.18 & 44.01 $\pm$ 0.18 & 32.03 $\pm$ 0.16 & 20.04 $\pm$ 0.13 \\
Vanilla ProtoNet~\cite{snell2017prototypical} & \textbf{73.65 $\pm$ 0.14} & 71.80 $\pm$ 0.15 & 69.19 $\pm$ 0.15 & 65.28 $\pm$ 0.16 & 59.52 $\pm$ 0.17 & 51.42 $\pm$ 0.18& 41.43 $\pm$ 0.18 & 32.29 $\pm$ 0.18 \\
RNNP~\cite{mazumder2021rnnp} & 73.47 $\pm$ 0.14 & 71.80 $\pm$ 0.15 & 69.37 $\pm$ 0.16 & 65.88 $\pm$ 0.17 & 60.51 $\pm$ 0.18 & 52.25 $\pm$ 0.19 & 41.74 $\pm$ 0.19 & 32.47 $\pm$ 0.19 \\
\hline
Median             & 73.54 $\pm$ 0.14 & 71.90 $\pm$ 0.15 & 69.30 $\pm$ 0.15 & 65.59 $\pm$ 0.16 & 59.88 $\pm$ 0.17 & 51.42 $\pm$ 0.18 & 41.13 $\pm$ 0.19 & 31.99 $\pm$ 0.18 \\
Absolute $T=10.0$  & 71.12 $\pm$ 0.15 & 69.58 $\pm$ 0.16 & 66.77 $\pm$ 0.17 & 62.27 $\pm$ 0.18 & 54.91 $\pm$ 0.20 & 45.13 $\pm$ 0.21 & 35.05 $\pm$ 0.20 & 28.20 $\pm$ 0.18 \\
Absolute $T=25.0$  & 73.10 $\pm$ 0.14 & 71.66 $\pm$ 0.15 & 69.13 $\pm$ 0.16 & 65.15 $\pm$ 0.17 & 58.63 $\pm$ 0.18 & 49.02 $\pm$ 0.19 & 38.40 $\pm$ 0.20 & 30.05 $\pm$ 0.18 \\
Absolute $T=50.0$  & 73.49 $\pm$ 0.14 & 71.88 $\pm$ 0.15 & 69.42 $\pm$ 0.16 & 65.52 $\pm$ 0.16 & 59.54 $\pm$ 0.18 & 50.65 $\pm$ 0.19 & 40.04 $\pm$ 0.19 & 31.34 $\pm$ 0.18 \\
Euclidean $T=10.0$ & 73.11 $\pm$ 0.15 & 71.60 $\pm$ 0.15 & 69.28 $\pm$ 0.16 & 65.57 $\pm$ 0.17 & 59.59 $\pm$ 0.18 & 50.45 $\pm$ 0.19 & 39.73 $\pm$ 0.19 & 30.88 $\pm$ 0.18 \\
Euclidean $T=25.0$ & 73.57 $\pm$ 0.14 & 71.98 $\pm$ 0.15 & 69.50 $\pm$ 0.16 & 65.78 $\pm$ 0.16 & 60.02 $\pm$ 0.18 & 51.59 $\pm$ 0.18 & 40.96 $\pm$ 0.19 & 31.98 $\pm$ 0.18 \\
Euclidean $T=50.0$ & 73.64 $\pm$ 0.14 & 71.96 $\pm$ 0.15 & 69.36 $\pm$ 0.16 & 65.68 $\pm$ 0.16 & 59.95 $\pm$ 0.17 & 51.58 $\pm$ 0.18 & 41.30 $\pm$ 0.19 & 32.18 $\pm$ 0.18 \\
Cosine $T=0.2$     & 73.62 $\pm$ 0.14 & 71.94 $\pm$ 0.15 & 69.44 $\pm$ 0.15 & 65.65 $\pm$ 0.16 & 59.91 $\pm$ 0.17 & 51.49 $\pm$ 0.18 & 41.14 $\pm$ 0.19 & 32.13 $\pm$ 0.18 \\
Cosine $T=0.5$     & 73.60 $\pm$ 0.14 & 71.85 $\pm$ 0.15 & 69.26 $\pm$ 0.15 & 65.46 $\pm$ 0.16 & 59.64 $\pm$ 0.17 & 51.50 $\pm$ 0.18 & 41.23 $\pm$ 0.19 & 32.18 $\pm$ 0.18 \\
Cosine $T=1.0$     & 73.57 $\pm$ 0.14 & 71.78 $\pm$ 0.15 & 69.13 $\pm$ 0.15 & 65.36 $\pm$ 0.16 & 59.62 $\pm$ 0.17 & 51.56 $\pm$ 0.18 & 41.44 $\pm$ 0.18 & 32.24 $\pm$ 0.18 \\
Cosine $T=2.0$     & \textbf{73.65 $\pm$ 0.14} & 71.83 $\pm$ 0.15 & 69.08 $\pm$ 0.16 & 65.25 $\pm$ 0.16 & 59.58 $\pm$ 0.17 & 51.26 $\pm$ 0.18 & 41.36 $\pm$ 0.18 & 32.10 $\pm$ 0.18 \\
Cosine $T=5.0$     & 73.55 $\pm$ 0.14 & 71.73 $\pm$ 0.15 & 69.06 $\pm$ 0.15 & 65.19 $\pm$ 0.16 & 59.42 $\pm$ 0.17 & 51.38 $\pm$ 0.18 & 41.31 $\pm$ 0.19 & 32.19 $\pm$ 0.18 \\
\toprule[0.8pt]		
TraNFS-$2$	       & 72.80 $\pm$ 0.15 & 71.86 $\pm$ 0.15 & 70.54 $\pm$ 0.16 & 68.25 $\pm$ 0.17 & 64.29 $\pm$ 0.19 & 57.04 $\pm$ 0.21 & \textbf{45.84 $\pm$ 0.24} & 35.09 $\pm$ 0.23 \\
TraNFS-$3$         & 73.17 $\pm$ 0.15 & \textbf{72.14 $\pm$ 0.15} & \textbf{70.71 $\pm$ 0.16} & \textbf{68.48 $\pm$ 0.17} & \textbf{64.59 $\pm$ 0.18} & \textbf{57.45 $\pm$ 0.21} & 45.80 $\pm$ 0.24 & \textbf{35.12 $\pm$ 0.23} \\
\bottomrule[1.2pt]
\end{tabular}
}
\label{tab:sym_swap_5w10k}
\end{table*}

\subsection{3-shot MiniImageNet}
We show 5-way 3-shot performance on MiniImageNet with symmetric label swap (Table~\ref{tab:sym_swap_5w3k}) and outlier (Table~\ref{tab:outlier_5w3k}) noise.
Note that we do not show results for paired label swap noise, as at $33.3\%$ noise, paired label noise is identical to symmetric, and at $66.7\%$, the clean class is dominated by the noisy class.

We observe similar trends as in the 5-way 5-shot experiments reported in Tables~\ref{tab:sym_swap_5w5k}, \ref{tab:pair_swap_5w5k}, and \ref{tab:outlier_5w5k}.
The baseline methods suffer dramatically from replacing a clean sample in the support set with a single noisy sample, with ProtoNet~\cite{snell2017prototypical} suffering almost a $14\%$ drop in accuracy in the $33.3\%$ symmetric label swap noise setting, as compared to the $5.71\%$ drop in accuracy from removing a shot.
Our proposed ProtoNet variants at various temperatures $T$ all outperform vanilla ProtoNet.
On the other hand, our TraNFS surpasses vanilla ProtoNet by $5.06\%$ and impressively is only $3.05\%$ short of the Oracle, despite not having knowledge of the noisy samples within the support set.

\begin{table}[t!]
\centering
\caption{
Few-shot performance with paired label swap noise on 5-way 10-shot MiniImageNet~\cite{vinyals2016matching}. 
}
\resizebox{\columnwidth}{!}{
\begin{tabular}{c || c c c} 
\toprule[1.2pt]
Model \textbackslash ~ Noise Proportion & $20\%$ & $30\%$ & $40\%$ \\
\toprule[0.8pt]		
Oracle             & 71.78 $\pm$ 0.15 & 70.82 $\pm$ 0.15 & 69.27 $\pm$ 0.16 \\
\hline
Nearest $k=1$      & 44.85 $\pm$ 0.18 & 40.80 $\pm$ 0.18 & 36.58 $\pm$ 0.17 \\
Nearest $k=3$      & 46.96 $\pm$ 0.18 & 42.31 $\pm$ 0.17 & 37.32 $\pm$ 0.16 \\
Nearest $k=5$      & 49.88 $\pm$ 0.18 & 45.21 $\pm$ 0.17 & 39.47 $\pm$ 0.17 \\
Linear Classifier  & 63.54 $\pm$ 0.16 & 56.70 $\pm$ 0.16 & 47.85 $\pm$ 0.16 \\
Matching Networks~\cite{vinyals2016matching} & 57.74 $\pm$ 0.19 & 52.80 $\pm$ 0.18 & 45.37 $\pm$ 0.17 \\
MAML~\cite{finn2017model} & 55.05 $\pm$ 0.18 & 41.95 $\pm$ 0.18 & 41.83 $\pm$ 0.18 \\
Vanilla ProtoNet~\cite{snell2017prototypical} & 68.34 $\pm$ 0.16 & 62.59 $\pm$ 0.16 & 52.73 $\pm$ 0.17 \\
RNNP~\cite{mazumder2021rnnp} & 68.89 $\pm$ 0.16 & 63.86 $\pm$ 0.17 & 54.06 $\pm$ 0.18 \\
\hline
Median             & 69.04 $\pm$ 0.15 & 63.50 $\pm$ 0.16 & 53.61 $\pm$ 0.17 \\
Absolute $T=50.0$  & 69.07 $\pm$ 0.16 & 63.62 $\pm$ 0.17 & 53.78 $\pm$ 0.18 \\
Absolute $T=25.0$  & 69.00 $\pm$ 0.16 & 63.75 $\pm$ 0.17 & 53.88 $\pm$ 0.18 \\
Absolute $T=10.0$  & 66.94 $\pm$ 0.17 & 61.82 $\pm$ 0.18 & 52.28 $\pm$ 0.20 \\
Euclidean $T=50.0$ & 68.86 $\pm$ 0.16 & 63.20 $\pm$ 0.17 & 53.24 $\pm$ 0.17 \\
Euclidean $T=25.0$ & 69.12 $\pm$ 0.16 & 63.48 $\pm$ 0.17 & 53.58 $\pm$ 0.17 \\
Euclidean $T=10.0$ & 68.91 $\pm$ 0.16 & 63.73 $\pm$ 0.17 & 53.81 $\pm$ 0.18 \\
Cosine $T=5.0$     & 68.50 $\pm$ 0.15 & 62.72 $\pm$ 0.16 & 52.76 $\pm$ 0.17 \\
Cosine $T=2.0$     & 68.42 $\pm$ 0.15 & 62.63 $\pm$ 0.16 & 52.87 $\pm$ 0.17 \\
Cosine $T=1.0$     & 68.48 $\pm$ 0.16 & 62.59 $\pm$ 0.17 & 52.86 $\pm$ 0.17 \\
Cosine $T=0.5$     & 68.58 $\pm$ 0.15 & 62.76 $\pm$ 0.16 & 52.90 $\pm$ 0.17 \\
Cosine $T=0.2$     & 68.82 $\pm$ 0.16 & 63.14 $\pm$ 0.17 & 53.27 $\pm$ 0.17 \\
\toprule[0.8pt]		
TraNFS-$2$	 & 70.13 $\pm$ 0.16 & 66.20 $\pm$ 0.17 & 56.97 $\pm$ 0.20 \\
TraNFS-$3$   & \textbf{70.38 $\pm$ 0.16} & \textbf{67.03 $\pm$ 0.18} & \textbf{58.94 $\pm$ 0.21} \\
\bottomrule[1.2pt]
\end{tabular}
}
\vspace{-4mm}
\label{tab:pair_swap_5w10k}
\end{table}

\begin{table*}[t!]
\centering
\setlength{\tabcolsep}{6pt}
\def\arraystretch{1.05}
\caption{
Few-shot performance with outlier noise on 5-way 10-shot MiniImageNet~\cite{vinyals2016matching}.
}
\resizebox{\textwidth}{!}{
\begin{tabular}{c || c c c c c c c c} 
\toprule[1.2pt]
Model \textbackslash ~ Noise Proportion & $0\%$ & $10\%$ & $20\%$ & $30\%$ & $40\%$ & $50\%$ & $60\%$ & $70\%$ \\
\toprule[0.8pt]		
Oracle             & 73.62 $\pm$ 0.14 & 72.78 $\pm$ 0.15 & 71.78 $\pm$ 0.15 & 70.82 $\pm$ 0.15 & 69.27 $\pm$ 0.16 & 64.70 $\pm$ 0.17 & 60.59 $\pm$ 0.17 & 53.88 $\pm$ 0.18 \\
\hline
Nearest $k=1$      & 53.14 $\pm$ 0.19 & 50.61 $\pm$ 0.19 & 48.25 $\pm$ 0.18 & 45.62 $\pm$ 0.18 & 42.91 $\pm$ 0.18 & 39.99 $\pm$ 0.17 & 37.06 $\pm$ 0.17 & 33.66 $\pm$ 0.17 \\
Nearest $k=3$      & 53.55 $\pm$ 0.19 & 51.49 $\pm$ 0.18 & 49.16 $\pm$ 0.18 & 46.50 $\pm$ 0.18 & 43.69 $\pm$ 0.17 & 40.63 $\pm$ 0.17 & 37.07 $\pm$ 0.16 & 33.15 $\pm$ 0.15 \\
Nearest $k=5$      & 54.81 $\pm$ 0.20 & 53.31 $\pm$ 0.19 & 51.46 $\pm$ 0.19 & 49.18 $\pm$ 0.18 & 46.35 $\pm$ 0.18 & 43.13 $\pm$ 0.17 & 39.32 $\pm$ 0.16 & 35.05 $\pm$ 0.16 \\
Linear Classifier  & 71.90 $\pm$ 0.15 & 69.62 $\pm$ 0.15 & 66.94 $\pm$ 0.16 & 63.70 $\pm$ 0.16 & 59.86 $\pm$ 0.16 & 55.31 $\pm$ 0.17 & 49.84 $\pm$ 0.17 & 43.42 $\pm$ 0.17 \\
Matching Networks~\cite{vinyals2016matching} & 62.68 $\pm$ 0.19 & 61.37 $\pm$ 0.19 & 59.58 $\pm$ 0.19 & 57.52 $\pm$ 0.19 & 54.58 $\pm$ 0.19 & 51.12 $\pm$ 0.19 & 46.48 $\pm$ 0.19 & 40.68 $\pm$ 0.18 \\
MAML~\cite{finn2017model} & 64.30 $\pm$ 0.18 & 64.43 $\pm$ 0.18 & 58.82 $\pm$ 0.18 & 51.30 $\pm$ 0.19 & 51.37 $\pm$ 0.19 & 51.36 $\pm$ 0.19 & 42.05 $\pm$ 0.19 & 30.89 $\pm$ 0.18 \\
Vanilla ProtoNet~\cite{snell2017prototypical} & 73.67 $\pm$ 0.14 & 72.27 $\pm$ 0.15 & 70.55 $\pm$ 0.15 & 68.08 $\pm$ 0.16 & 64.93 $\pm$ 0.16 & 60.66 $\pm$ 0.17 & 55.28 $\pm$ 0.18 & 47.94 $\pm$ 0.19 \\
RNNP~\cite{mazumder2021rnnp} & 73.35 $\pm$ 0.14 & 71.92 $\pm$ 0.15 & 70.16 $\pm$ 0.15 & 67.97 $\pm$ 0.16 & 64.90 $\pm$ 0.17 & 60.81 $\pm$ 0.17 & 55.34 $\pm$ 0.18 & 48.07 $\pm$ 0.19 \\
\hline
Median             & \textbf{73.69 $\pm$ 0.14} & \textbf{72.50 $\pm$ 0.15} & 70.78 $\pm$ 0.15 & 68.47 $\pm$ 0.15 & 65.26 $\pm$ 0.16 & 61.07 $\pm$ 0.17 & 55.46 $\pm$ 0.18 & 47.92 $\pm$ 0.19 \\
Absolute $T=50.0$  & 73.56 $\pm$ 0.14 & 72.44 $\pm$ 0.15 & 70.82 $\pm$ 0.15 & 68.60 $\pm$ 0.16 & 65.48 $\pm$ 0.16 & 61.33 $\pm$ 0.17 & 55.62 $\pm$ 0.18 & 48.19 $\pm$ 0.19 \\
Absolute $T=25.0$  & 73.26 $\pm$ 0.14 & 72.14 $\pm$ 0.15 & 70.65 $\pm$ 0.15 & 68.57 $\pm$ 0.16 & 65.53 $\pm$ 0.17 & 61.29 $\pm$ 0.17 & 55.45 $\pm$ 0.18 & 47.89 $\pm$ 0.19 \\
Absolute $T=10.0$  & 71.10 $\pm$ 0.15 & 69.96 $\pm$ 0.15 & 68.48 $\pm$ 0.16 & 66.29 $\pm$ 0.17 & 63.36 $\pm$ 0.17 & 58.83 $\pm$ 0.18 & 52.58 $\pm$ 0.19 & 44.80 $\pm$ 0.20 \\
Euclidean $T=50.0$ & 73.62 $\pm$ 0.14 & 72.39 $\pm$ 0.15 & 70.59 $\pm$ 0.15 & 68.28 $\pm$ 0.16 & 65.21 $\pm$ 0.16 & 60.94 $\pm$ 0.17 & 55.37 $\pm$ 0.18 & 48.04 $\pm$ 0.19 \\
Euclidean $T=25.0$ & 73.58 $\pm$ 0.14 & 72.36 $\pm$ 0.15 & 70.70 $\pm$ 0.15 & 68.40 $\pm$ 0.16 & 65.33 $\pm$ 0.16 & 61.08 $\pm$ 0.17 & 55.15 $\pm$ 0.18 & 47.79 $\pm$ 0.19 \\
Euclidean $T=10.0$ & 73.19 $\pm$ 0.15 & 72.03 $\pm$ 0.15 & 70.48 $\pm$ 0.16 & 68.21 $\pm$ 0.16 & 65.00 $\pm$ 0.17 & 60.50 $\pm$ 0.18 & 54.51 $\pm$ 0.19 & 46.58 $\pm$ 0.20 \\
Cosine $T=5.0$     & 73.57 $\pm$ 0.14 & 72.25 $\pm$ 0.15 & 70.44 $\pm$ 0.15 & 67.97 $\pm$ 0.16 & 64.77 $\pm$ 0.16 & 60.61 $\pm$ 0.17 & 55.14 $\pm$ 0.18 & 47.94 $\pm$ 0.19 \\
Cosine $T=2.0$     & 73.63 $\pm$ 0.14 & 72.28 $\pm$ 0.14 & 70.47 $\pm$ 0.15 & 68.10 $\pm$ 0.16 & 64.79 $\pm$ 0.16 & 60.60 $\pm$ 0.17 & 55.02 $\pm$ 0.18 & 48.03 $\pm$ 0.19 \\
Cosine $T=1.0$     & 73.46 $\pm$ 0.14 & 72.19 $\pm$ 0.15 & 70.33 $\pm$ 0.15 & 67.97 $\pm$ 0.16 & 64.80 $\pm$ 0.16 & 60.59 $\pm$ 0.17 & 55.09 $\pm$ 0.18 & 47.88 $\pm$ 0.19 \\
Cosine $T=0.5$     & 73.64 $\pm$ 0.14 & 72.30 $\pm$ 0.15 & 70.53 $\pm$ 0.15 & 68.13 $\pm$ 0.16 & 65.07 $\pm$ 0.16 & 60.73 $\pm$ 0.17 & 55.16 $\pm$ 0.18 & 48.13 $\pm$ 0.19 \\
Cosine $T=0.2$     & 73.55 $\pm$ 0.14 & 72.40 $\pm$ 0.15 & 70.61 $\pm$ 0.15 & 68.37 $\pm$ 0.16 & 65.26 $\pm$ 0.16 & 61.00 $\pm$ 0.17 & 55.34 $\pm$ 0.18 & 47.97 $\pm$ 0.19 \\
\toprule[0.8pt]		
TraNFS-$2$	       & 72.43 $\pm$ 0.15 & 71.54 $\pm$ 0.16 & 70.24 $\pm$ 0.16 & 68.56 $\pm$ 0.17 & 65.93 $\pm$ 0.18 & 62.21 $\pm$ 0.20 & 56.98 $\pm$ 0.21 & 49.41 $\pm$ 0.22 \\
TraNFS-$3$         & 72.91 $\pm$ 0.15 & 72.12 $\pm$ 0.15 & \textbf{70.92 $\pm$ 0.16} & \textbf{69.47 $\pm$ 0.16} & \textbf{67.14 $\pm$ 0.17} & \textbf{63.60 $\pm$ 0.19} & \textbf{58.68 $\pm$ 0.20} & \textbf{50.66 $\pm$ 0.22} \\
\bottomrule[1.2pt]
\end{tabular}
}
\label{tab:outlier_5w10k}
\end{table*}

\begin{figure*}[ht]
  \centering
  \hfill
  \begin{subfigure}{0.47\linewidth}
    \centering
    \includegraphics[width=1.0\textwidth]{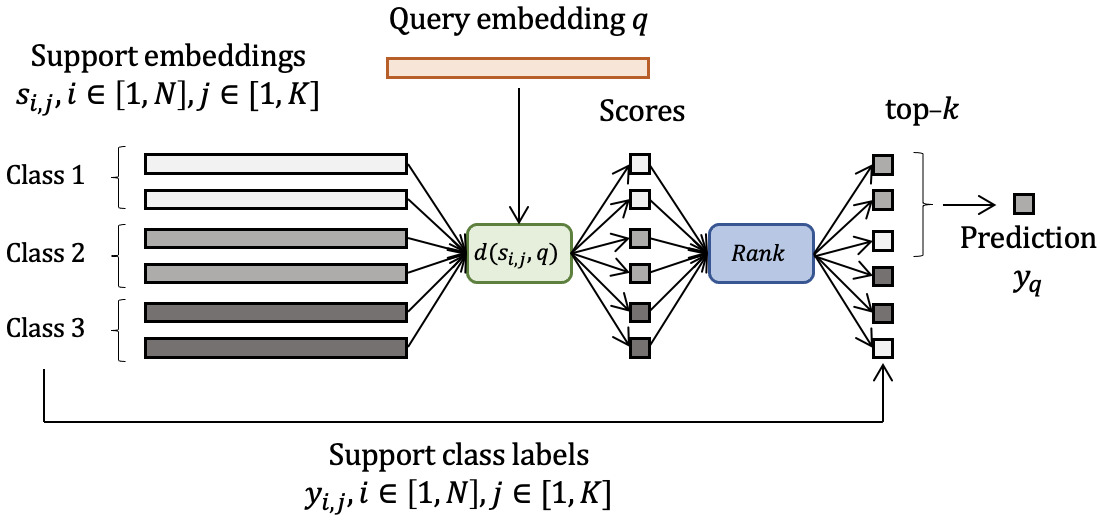}
    \caption{Nearest Neighbors}
    \label{fig:methods_nn}
  \end{subfigure}
  \hfill
  \begin{subfigure}{0.47\linewidth}
    \centering
    \includegraphics[width=1.0\textwidth]{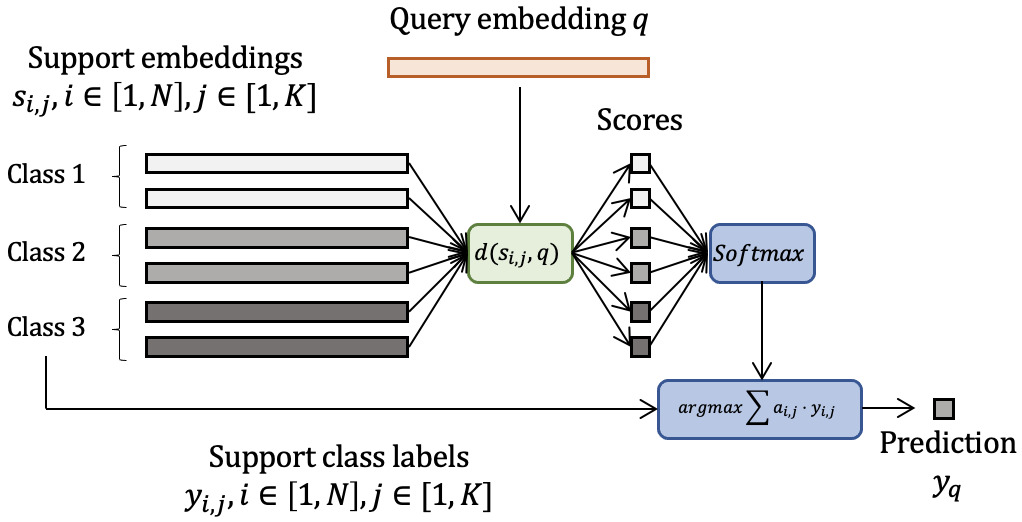}
    \caption{Matching Networks}
    \label{fig:methods_match}
  \end{subfigure}
  \hfill
  \\
  \hfill
  \begin{subfigure}{0.35\linewidth}
    \centering
    \includegraphics[width=1.0\textwidth]{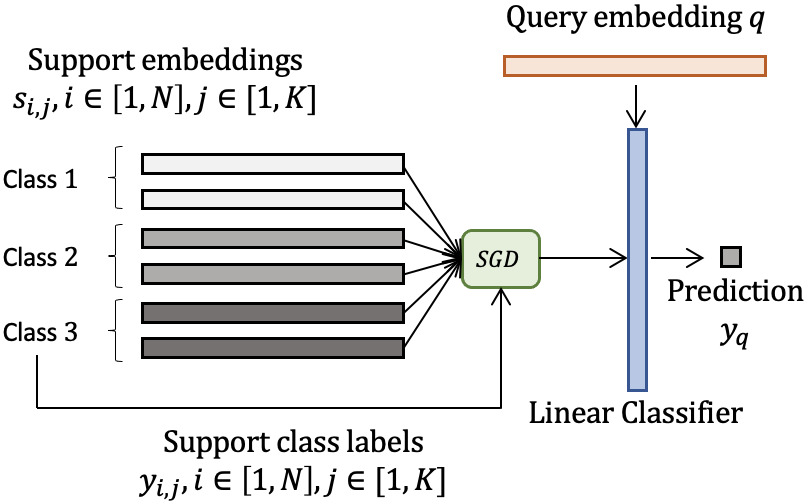}
    \caption{Linear Classifier}
    \label{fig:methods_lin_cls}
  \end{subfigure}    
  \hfill
  \begin{subfigure}{0.63\linewidth}
    \centering
    \includegraphics[width=1.0\textwidth]{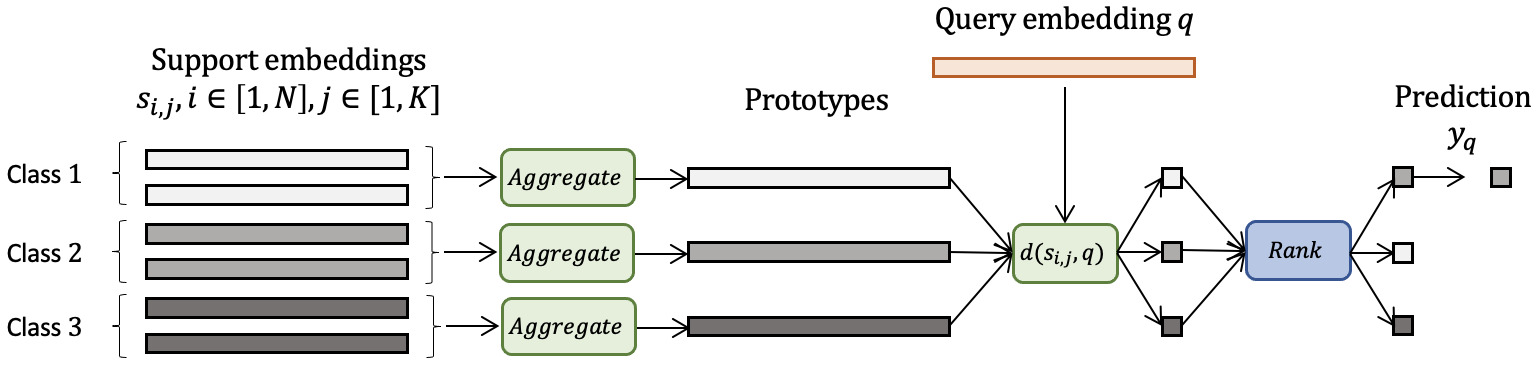}
    \caption{Prototypical Networks}
    \label{fig:methods_proto}
  \end{subfigure}
  \hfill
\caption{Visual overview of several of the few-shot method archetypes considered.}
\label{fig:methods}
\end{figure*}

\subsection{10-shot MiniImageNet}
We show 5-way 10-shot performance on MiniImageNet with symmetric label swap (Table~\ref{tab:sym_swap_5w10k}), paired label swap (Table~\ref{tab:pair_swap_5w10k}), and outlier (Table~\ref{tab:outlier_5w10k}) noise.
Note that we only show $20\%$, $30\%$, and $40\%$ noise proportion for paired label swap noise, as at $0\%$ and $10\%$, paired label swapping is no different from symmetric swapping (Table~\ref{tab:sym_swap_5w10k}) for 10 shots, and at $50\%$ and above the noisy class would have either a share of or the outright majority.

Our proposed TraNFS shines with 10-shot tasks as well.
As in the 5-shot case, our method does especially well in moderate to high noise levels.
In particular, we observe over $5\%$ absolute improvement from TraNFS over vanilla ProtoNet at $40\%$ and $50\%$ symmetric label swap noise and an impressive $6.21\%$ improvement for $40\%$ paired label swap noise. 
TraNFS is also the best method for rejecting outlier noise as well.

\section{Method descriptions}
\label{apx:baselines}
Fig.~\ref{fig:methods} shows a visual comparison of some of the baselines we compare against. We discuss implementation details below.

\minisection{Oracle} 
When noise appears in a support set, accuracy of the few-shot model is reduced for two reasons: (1) mislabeled samples provide the model with misleading information about the class, and (2) clean samples that would have otherwise been informative were removed from the support set.
FSL performance can be heavily influenced by the number of shots, especially in the low-data regime, so we find it important to separate out the aforementioned two sources of performance degradation.
For this purpose, we include in our results tables an \textit{Oracle} model consisting of a vanilla ProtoNet~\cite{snell2017prototypical} with prototypes produced from only the correctly labeled samples in the support set.
Note that the Oracle requires knowing the identities of the noisy samples, which cannot be reasonably expected in many real-world settings and is thus not a fair comparison with the other methods, but we include it to give a sense of constructive information content still available in the support set after noise corruption.

\minisection{Nearest Neighbors}
In the context of FSL, nearest neighbors (Fig.~\ref{fig:methods_nn}) is a simple, non-parametric classification technique which classifies query samples based on the labels of the $k$ closest support samples in embedding space.
Whichever class has the plurality among the $k$ nearest neighbor support samples is the prediction, with ties broken uniformly at random among the tied classes.
We report results for $k\in\{1,3,5\}$.

\minisection{Linear Classifier}
We train a single fully connected layer $\mathbb{R}^D \to \mathbb{R}^N$ on top of frozen convolutional features (Fig.~\ref{fig:methods_lin_cls}).
For each episode, the parameters of the fully connected layer are learned with the AdamW~\cite{loshchilov2018decoupled} optimizer with weight decay $0.01$, trained for $100$ steps.
Note that this approach resembles the Baseline method~\cite{chen2019closer}, with the primary difference being that we use the ProtoNet objective and episodic meta-training to learn the feature extractor $\mathcal F$, as opposed to the softmax cross entropy loss with batch learning on the base classes.

\minisection{Matching Networks~\cite{vinyals2016matching}}
Matching networks (Fig.~\ref{fig:methods_match}) use an attention mechanism to compare the embedded query sample with embeddings of each of the support set samples, with the prediction being a linear combination of the support set labels based on the result of this attention.
While this mechanism is trainable in a meta-learning setup, we found that we achieved better results than those reported in the literature by using a frozen convolutional feature extractor trained with the ProtoNet loss.

\minisection{MAML~\cite{finn2017model}}
Model-Agnostic Meta-Learning (MAML) seeks to learn a good initialization so that the model can be quickly adapted to new tasks, with this initialization learned through second-order gradients.
As such, unlike the other methods we compare against, we do not use the weights of the same frozen 4-layer convolutional feature extractor for MAML.
Instead, we use the Adam optimizer to train MAML with a meta-learning rate of $3\times 10^{-3}$ and inner loop learning rate of $1\times 10^{-2}$, using $5$ adaptation steps during meta-training and $10$ steps during meta-test.
We use the same random horizontal flips, resized crops, and color jitters for data augmentations as the rest of our experiments.

\begin{table*}[t!]
\centering
\caption{
{\bf Temperature sweep for our ProtoNet variants: symmetric label swap noise}. 5-way 5-shot Acc. $\pm$ 95\% CI on {\textcolor{blue}{MiniImageNet}}~\cite{vinyals2016matching}, {\textcolor{darkgreen}{TieredImageNet}}~\cite{ren2018meta}.
Best viewed in color.
}
\vspace{-3mm}
\setlength{\tabcolsep}{6pt}
\def\arraystretch{1.05}
\resizebox{1.00\textwidth}{!}{
\begin{tabular}{c || j k j k j k j k} 
    \toprule[1.2pt]
Model \textbackslash ~ Noise Proportion & \multicolumn{2}{c}{$0\%$}  & \multicolumn{2}{c}{$20\%$} & \multicolumn{2}{c}{$40\%$}  & \multicolumn{2}{c}{$60\%$} \\
\toprule[0.8pt]		
\hline
Absolute $T=50.0$  & 68.18 $\pm$ 0.16 & 71.24 $\pm$ 0.18 & 62.98 $\pm$ 0.17 & 66.56 $\pm$ 0.20 & 51.68 $\pm$ 0.19 & 54.97 $\pm$ 0.21 & 39.24 $\pm$ 0.20 & 41.59 $\pm$ 0.21 \\
Absolute $T=25.0$  & 68.24 $\pm$ 0.16 & 71.27 $\pm$ 0.18 & \textbf{63.46 $\pm$ 0.17} & 66.87 $\pm$ 0.20 & 52.06 $\pm$ 0.20 & 55.26 $\pm$ 0.22 & 39.78 $\pm$ 0.20 & 42.54 $\pm$ 0.22 \\
Absolute $T=10.0$  & 67.15 $\pm$ 0.17 & 70.15 $\pm$ 0.19 & 62.96 $\pm$ 0.18 & 66.10 $\pm$ 0.20 & 52.08 $\pm$ 0.20 & 55.08 $\pm$ 0.23 & \textbf{39.92 $\pm$ 0.21} & 42.49 $\pm$ 0.23 \\
Absolute $T=5.0$  & 63.89 $\pm$ 0.17 & 66.56 $\pm$ 0.19 & 59.63 $\pm$ 0.18 & 62.67 $\pm$ 0.21 & 51.30 $\pm$ 0.20 & 53.83 $\pm$ 0.22 & 37.99 $\pm$ 0.21 & 39.91 $\pm$ 0.23 \\
Absolute $T=1.0$  & 50.26 $\pm$ 0.20 & 51.39 $\pm$ 0.22 & 47.04 $\pm$ 0.20 & 48.40 $\pm$ 0.23 & 40.40 $\pm$ 0.21 & 41.45 $\pm$ 0.23 & 31.03 $\pm$ 0.20 & 31.75 $\pm$ 0.21 \\
\hline
Euclidean $T=50.0$ & 68.31 $\pm$ 0.16 & 71.31 $\pm$ 0.18 & 62.78 $\pm$ 0.17 & 66.36 $\pm$ 0.19 & 51.86 $\pm$ 0.19 & 55.19 $\pm$ 0.21 & 38.90 $\pm$ 0.20 & 41.19 $\pm$ 0.21 \\
Euclidean $T=25.0$ & 68.32 $\pm$ 0.16 & \textbf{71.48 $\pm$ 0.18} & 63.02 $\pm$ 0.17 & 66.69 $\pm$ 0.19 & 52.09 $\pm$ 0.19 & 55.62 $\pm$ 0.21 & 39.33 $\pm$ 0.20 & 41.75 $\pm$ 0.21 \\
Euclidean $T=10.0$ & 68.23 $\pm$ 0.16 & 71.18 $\pm$ 0.19 & \textbf{63.46 $\pm$ 0.17} & \textbf{67.04 $\pm$ 0.20} & 52.24 $\pm$ 0.20 & 55.78 $\pm$ 0.22 & 39.87 $\pm$ 0.20 & 42.53 $\pm$ 0.22 \\
Euclidean $T=5.0$  & 67.53 $\pm$ 0.16 & 70.54 $\pm$ 0.18 & 63.00 $\pm$ 0.18 & 66.56 $\pm$ 0.20 & \textbf{53.79 $\pm$ 0.20} & \textbf{57.37 $\pm$ 0.22} & 39.63 $\pm$ 0.21 & 42.31 $\pm$ 0.22 \\
Euclidean $T=1.0$  & 56.75 $\pm$ 0.19 & 59.17 $\pm$ 0.21 & 52.31 $\pm$ 0.19 & 54.82 $\pm$ 0.22 & 44.06 $\pm$ 0.20 & 46.09 $\pm$ 0.23 & 32.88 $\pm$ 0.20 & 33.99 $\pm$ 0.21 \\
\hline
Cosine $T=10.0$    & 68.24 $\pm$ 0.16 & 71.27 $\pm$ 0.18 & 62.47 $\pm$ 0.17 & 66.16 $\pm$ 0.19 & 51.41 $\pm$ 0.19 & 54.96 $\pm$ 0.21 & 38.38 $\pm$ 0.19 & 40.74 $\pm$ 0.21 \\
Cosine $T=5.0$     & 68.31 $\pm$ 0.16 & 71.16 $\pm$ 0.18 & 62.51 $\pm$ 0.17 & 65.99 $\pm$ 0.20 & 51.51 $\pm$ 0.19 & 54.78 $\pm$ 0.21 & 38.55 $\pm$ 0.19 & 40.81 $\pm$ 0.21 \\
Cosine $T=2.0$     & 68.28 $\pm$ 0.16 & 71.22 $\pm$ 0.18 & 62.57 $\pm$ 0.17 & 66.24 $\pm$ 0.19 & 51.59 $\pm$ 0.19 & 55.06 $\pm$ 0.21 & 38.71 $\pm$ 0.19 & 40.99 $\pm$ 0.21 \\
Cosine $T=1.0$     & 68.21 $\pm$ 0.16 & 71.21 $\pm$ 0.18 & 62.70 $\pm$ 0.17 & 66.47 $\pm$ 0.19 & 51.72 $\pm$ 0.19 & 55.27 $\pm$ 0.21 & 38.92 $\pm$ 0.19 & 41.32 $\pm$ 0.21 \\
Cosine $T=0.5$     & \textbf{68.42 $\pm$ 0.16} & 71.31 $\pm$ 0.18 & 63.13 $\pm$ 0.18 & 66.81 $\pm$ 0.20 & 52.08 $\pm$ 0.19 & 55.60 $\pm$ 0.22 & 39.36 $\pm$ 0.20 & 42.14 $\pm$ 0.22 \\
Cosine $T=0.2$     & 68.20 $\pm$ 0.16 & 70.59 $\pm$ 0.18 & \textbf{63.46 $\pm$ 0.17} & 66.62 $\pm$ 0.20 & 52.42 $\pm$ 0.20 & 55.78 $\pm$ 0.22 & 39.90 $\pm$ 0.20 & \textbf{42.56 $\pm$ 0.22} \\
Cosine $T=0.1$     & 67.52 $\pm$ 0.16 & 69.30 $\pm$ 0.19 & 63.07 $\pm$ 0.18 & 65.25 $\pm$ 0.20 & 52.22 $\pm$ 0.20 & 54.24 $\pm$ 0.23 & 39.85 $\pm$ 0.21 & 41.79 $\pm$ 0.23 \\
\bottomrule[1.2pt]
\end{tabular}
}
\vspace{-3mm}
\label{tab:sym_swap_5w5k_T_sweep}
\end{table*}

\begin{table}[t]
\centering
\caption{
{\bf Temperature sweep for our ProtoNet variants: paired label swap noise}. 5-way 5-shot Acc. $\pm$ 95\% CI on {\textcolor{blue}{MiniImageNet}}~\cite{vinyals2016matching}, {\textcolor{darkgreen}{TieredImageNet}}~\cite{ren2018meta}.
Best viewed in color.
}
\vspace{-3mm}
\resizebox{0.85\columnwidth}{!}{
\begin{tabular}{c || j k } 
\toprule[1.2pt]
Model \textbackslash ~ Noise Proportion & \multicolumn{2}{c}{$40\%$} \\
\toprule[0.8pt]		
Absolute $T=50.0$  & 48.64 $\pm$ 0.19 & 51.83 $\pm$ 0.21 \\
Absolute $T=25.0$  & 49.38 $\pm$ 0.20 & 52.40 $\pm$ 0.22 \\
Absolute $T=10.0$  & 49.56 $\pm$ 0.20 & 52.54 $\pm$ 0.23 \\
Absolute $T=5.0$   & 47.18 $\pm$ 0.21 & 49.42 $\pm$ 0.23 \\
Absolute $T=1.0$   & 37.85 $\pm$ 0.21 & 38.47 $\pm$ 0.23 \\
\hline
Euclidean $T=50.0$ & 48.43 $\pm$ 0.19 & 51.39 $\pm$ 0.21 \\
Euclidean $T=25.0$ & 48.67 $\pm$ 0.19 & 51.90 $\pm$ 0.21 \\
Euclidean $T=10.0$ & 49.37 $\pm$ 0.19 & 52.55 $\pm$ 0.22 \\
Euclidean $T=5.0$  & \textbf{49.75 $\pm$ 0.20} & 52.57 $\pm$ 0.22 \\
Euclidean $T=1.0$  & 41.30 $\pm$ 0.21 & 42.92 $\pm$ 0.22 \\
\hline
Cosine $T=10.0$    & 47.75 $\pm$ 0.19 & 50.95 $\pm$ 0.21  \\
Cosine $T=5.0$     & 48.03 $\pm$ 0.19 & 51.17 $\pm$ 0.21  \\
Cosine $T=2.0$     & 48.03 $\pm$ 0.19 & 51.19 $\pm$ 0.21  \\
Cosine $T=1.0$     & 48.53 $\pm$ 0.19 & 51.71 $\pm$ 0.21  \\
Cosine $T=0.5$     & 48.90 $\pm$ 0.19 & 52.14 $\pm$ 0.21  \\
Cosine $T=0.2$     & 49.40 $\pm$ 0.19 & \textbf{52.72 $\pm$ 0.22}  \\
Cosine $T=0.1$     & 49.71 $\pm$ 0.20 & 51.96 $\pm$ 0.23  \\
\bottomrule[1.2pt]
\end{tabular}
}
\vspace{-4mm}
\label{tab:pair_swap_5w5k_T_sweep}
\end{table}

\begin{table*}[t!]
\centering
\setlength{\tabcolsep}{6pt}
\def\arraystretch{1.05}
\caption{
{\bf Temperature sweep for our ProtoNet variants: outlier noise}. 5-way 5-shot Acc. $\pm$ 95\% CI on {\textcolor{blue}{MiniImageNet}}~\cite{vinyals2016matching}, {\textcolor{darkgreen}{TieredImageNet}}~\cite{ren2018meta}.
Best viewed in color.
}
\vspace{-3mm}
\resizebox{1.0\textwidth}{!}{
\begin{tabular}{c || j k j k j k j k} 
    \toprule[1.2pt]
Model \textbackslash ~ Noise Proportion & \multicolumn{2}{c}{$0\%$} & \multicolumn{2}{c}{$20\%$} & \multicolumn{2}{c}{$40\%$} & \multicolumn{2}{c}{$60\%$} \\
\toprule[0.8pt]		
Absolute $T=50.0$  & 68.41 $\pm$ 0.16 & 71.42 $\pm$ 0.19 & 64.62 $\pm$ 0.17 & 67.96 $\pm$ 0.19 & 58.08 $\pm$ 0.19 & 61.68 $\pm$ 0.21 & 47.33 $\pm$ 0.20 & 50.71 $\pm$ 0.22 \\
Absolute $T=25.0$  & 68.13 $\pm$ 0.16 & 71.17 $\pm$ 0.18 & 64.69 $\pm$ 0.17 & 68.00 $\pm$ 0.19 & 58.30 $\pm$ 0.18 & 61.98 $\pm$ 0.21 & \textbf{47.39 $\pm$ 0.20} & 50.59 $\pm$ 0.22 \\
Absolute $T=10.0$  & 67.18 $\pm$ 0.16 & 70.10 $\pm$ 0.19 & 64.14 $\pm$ 0.17 & 67.29 $\pm$ 0.20 & 58.12 $\pm$ 0.19 & 61.65 $\pm$ 0.21 & 47.02 $\pm$ 0.21 & 49.68 $\pm$ 0.22 \\
Absolute $T=5.0$ & 63.97 $\pm$ 0.17 & 66.78 $\pm$ 0.19 & 60.96 $\pm$ 0.18 & 63.88 $\pm$ 0.20 & 55.24 $\pm$ 0.19 & 58.17 $\pm$ 0.21 & 44.28 $\pm$ 0.21 & 46.50 $\pm$ 0.22 \\
Absolute $T=1.0$ & 50.02 $\pm$ 0.19 & 51.77 $\pm$ 0.22 & 47.71 $\pm$ 0.20 & 49.01 $\pm$ 0.22 & 42.90 $\pm$ 0.20 & 44.45 $\pm$ 0.23 & 34.52 $\pm$ 0.20 & 35.08 $\pm$ 0.21 \\
\hline
Euclidean $T=50.0$ & 68.31 $\pm$ 0.16 & 71.14 $\pm$ 0.18 & 64.25 $\pm$ 0.17 & 67.53 $\pm$ 0.19 & 57.43 $\pm$ 0.18 & 60.95 $\pm$ 0.21 & 47.06 $\pm$ 0.20 & 50.34 $\pm$ 0.21 \\
Euclidean $T=25.0$ & \textbf{68.51 $\pm$ 0.16} & 71.28 $\pm$ 0.18 & 64.57 $\pm$ 0.17 & 67.89 $\pm$ 0.19 & 58.01 $\pm$ 0.18 & 61.61 $\pm$ 0.20 & 47.25 $\pm$ 0.20 & 50.49 $\pm$ 0.21 \\
Euclidean $T=10.0$ & 68.19 $\pm$ 0.16 & 71.20 $\pm$ 0.18 & 64.55 $\pm$ 0.17 & 68.02 $\pm$ 0.19 & 58.17 $\pm$ 0.19 & 62.00 $\pm$ 0.21 & 47.24 $\pm$ 0.20 & 50.86 $\pm$ 0.22 \\
Euclidean $T=5.0$  & 67.58 $\pm$ 0.16 & 70.45 $\pm$ 0.18 & 64.25 $\pm$ 0.17 & 67.55 $\pm$ 0.19 & 57.82 $\pm$ 0.19 & 61.69 $\pm$ 0.21 & 46.34 $\pm$ 0.20 & 50.21 $\pm$ 0.22 \\
Euclidean $T=1.0$  & 56.94 $\pm$ 0.18 & 59.04 $\pm$ 0.21 & 53.59 $\pm$ 0.19 & 55.57 $\pm$ 0.22 & 47.23 $\pm$ 0.20 & 49.63 $\pm$ 0.22 & 37.32 $\pm$ 0.20 & 39.37 $\pm$ 0.22 \\
\hline
Cosine $T=10.0$    & 68.41 $\pm$ 0.16 & 71.20 $\pm$ 0.18 & 64.19 $\pm$ 0.17 & 67.48 $\pm$ 0.19 & 57.33 $\pm$ 0.18 & 60.87 $\pm$ 0.21 & 47.02 $\pm$ 0.20 & 50.12 $\pm$ 0.21 \\
Cosine $T=5.0$     & 68.29 $\pm$ 0.16 & 71.28 $\pm$ 0.19 & 64.04 $\pm$ 0.17 & 67.46 $\pm$ 0.20 & 57.30 $\pm$ 0.18 & 61.10 $\pm$ 0.21 & 47.08 $\pm$ 0.20 & 50.27 $\pm$ 0.21 \\
Cosine $T=2.0$     & 68.29 $\pm$ 0.16 & 71.20 $\pm$ 0.18 & 64.13 $\pm$ 0.17 & 67.53 $\pm$ 0.19 & 57.39 $\pm$ 0.18 & 61.07 $\pm$ 0.20 & 46.97 $\pm$ 0.20 & 50.23 $\pm$ 0.21 \\
Cosine $T=1.0$     & 68.30 $\pm$ 0.16 & \textbf{71.54 $\pm$ 0.18} & 64.23 $\pm$ 0.17 & 68.07 $\pm$ 0.19 & 57.51 $\pm$ 0.18 & 61.82 $\pm$ 0.20 & 46.89 $\pm$ 0.20 & 50.82 $\pm$ 0.21 \\
Cosine $T=0.5$     & 68.35 $\pm$ 0.16 & 71.38 $\pm$ 0.18 & 64.51 $\pm$ 0.17 & \textbf{68.16 $\pm$ 0.19} & 57.97 $\pm$ 0.18 & 62.13 $\pm$ 0.20 & 47.28 $\pm$ 0.20 & \textbf{51.14 $\pm$ 0.22} \\
Cosine $T=0.2$     & 68.20 $\pm$ 0.16 & 70.79 $\pm$ 0.18 & \textbf{64.78 $\pm$ 0.17} & 67.94 $\pm$ 0.19 & 58.36 $\pm$ 0.18 & \textbf{62.37 $\pm$ 0.21} & 47.34 $\pm$ 0.20 & 51.12 $\pm$ 0.22 \\
Cosine $T=0.1$     & 67.82 $\pm$ 0.16 & 69.33 $\pm$ 0.19 & 64.49 $\pm$ 0.17 & 66.55 $\pm$ 0.20 & \textbf{58.42 $\pm$ 0.19} & 61.21 $\pm$ 0.21 & 46.90 $\pm$ 0.21 & 50.00 $\pm$ 0.22 \\
\bottomrule[1.2pt]
\end{tabular}
}
\vspace{-4mm}
\label{tab:outlier_5w5k_T_sweep}
\end{table*}

\minisection{ProtoNet~\cite{snell2017prototypical}}
ProtoNet (Fig.~\ref{fig:methods_proto}) was introduced in Sec.~\ref{sec:prelim}.
We refer to the version of ProtoNet proposed by Snell \etal in \cite{snell2017prototypical} (using the mean of the support embeddings) as \textit{Vanilla ProtoNet} to distinguish it from the median and similarity weighted variants of ProtoNet that we propose in Sec.~\ref{sec:static_agg}.

\minisection{Baseline++~\cite{chen2019closer}} 
Baseline++ was proposed as a simple alternative to recent few-shot methods.
Rather than requiring relatively complex bi-level meta-training, \cite{chen2019closer} proposed simply pre-training a feature extractor with a standard supervised cross-entropy loss, freezing the feature extractor's weights, and then fine-tuning a one-layer classifier just on top of the few examples in the novel class's support set features.
In particular, the Baseline++ method uses cosine similarity and a softmax for the classifier.
Such an approach has been shown to be surprisingly competitive with popular few-shot approaches.
We implement this cosine similarity classifier in our framework, with the primary difference being that we use a feature extractor trained with the ProtoNet loss instead of a cross-entropy loss, in order to compare the classifier design on even terms.
Note that \cite{chen2019closer} also proposed a simpler approach using a standard linear layer instead of cosine distance, which they referred to as Baseline; other than the training objective of the fixed feature extractor, the Baseline method is equivalent to our Linear Classifier baseline.

\minisection{NegMargin~\cite{liu2020negative}} Taking insights from the metric learning literature, \cite{liu2020negative} suggests that discriminability shortcomings of the softmax loss can be mitigated by learning with a margin.
Surprisingly, NegMargin found that positive margins underperform in open-set few-shot classification scenarios, while negative margins can lead to significant improvements in performance due to improved transferability.
To perform few-shot classification, NegMargin takes a similar approach to \cite{chen2019closer}--first pre-training and then freezing the feature extractor, followed by fine-tuning of a classifier for the novel support set--with the primary difference being the substitution of the standard softmax with the negative margin softmax loss during pre-training.
As such, unlike the other methods we compare against, we do not use the weights of the same frozen 4-layer convolutional feature extractor for NegMargin.
We use the official NegMargin codebase,\footnote{\url{https://github.com/bl0/negative-margin.few-shot}} modifying their code to inject artificial noisy labels into support sets during meta-test evaluation.

\minisection{RNNP~\cite{mazumder2021rnnp}}
Robust Nearest Neighbor Prototype (RNNP) creates hybrid examples by interpolating between samples within each support set, somewhat similarly to mixup.
Using ProtoNet prototypes of the original support embeddings as initialization for the class centers, $k$-means is then used to refine the prototypes in an unsupervised manner.
We reproduce RNNP, using the suggested $K-1$ hybrids per support sample and mixing ratio of $0.8$ when producing hybrids.

\section{Additional ablation studies}
\label{apx:trans_abl}

\begin{table}[t]
\centering
\caption{
{\bf Ablation study: Clean Prototype Loss} for a 3-layer TraNFS trained on 5-way 5-shot MiniImageNet~\cite{vinyals2016matching}.
}
\vspace{-3mm}
\resizebox{\columnwidth}{!}{
\begin{tabular}{c || c c c c} 
\toprule[1.2pt]
$\lambda_c$ & $0\%$ & $20\%$ & $40\%$ & $60\%$ \\
\toprule[0.8pt]		
0.0 & 63.77 $\pm$ 0.18 & 60.67 $\pm$ 0.19 & 53.14 $\pm$ 0.22 & 39.75 $\pm$ 0.23 \\
0.1 & 65.68 $\pm$ 0.18 & 61.94 $\pm$ 0.19 & 53.45 $\pm$ 0.22 & 39.20 $\pm$ 0.24 \\
0.5 & 68.11 $\pm$ 0.17 & 64.56 $\pm$ 0.18 & 56.47 $\pm$ 0.21 & 41.94 $\pm$ 0.24 \\
1.0 & \textbf{68.80 $\pm$ 0.16} & \textbf{65.10 $\pm$ 0.18} & \textbf{57.26 $\pm$ 0.21} & \textbf{42.82 $\pm$ 0.24} \\
5.0 & 68.53 $\pm$ 0.17 & 65.08 $\pm$ 0.18 & 56.65 $\pm$ 0.21 & 42.60 $\pm$ 0.24 \\
10.0 & 68.76 $\pm$ 0.17 & 64.87 $\pm$ 0.18 & 56.76 $\pm$ 0.21 & 42.17 $\pm$ 0.24 \\
\bottomrule[1.2pt]
\end{tabular}
}
\label{tab:clean_proto_abl}
\end{table}

\minisection{Feature extractor training objective} We consider the performance of few-shot learning methods within the context of support set noise primarily with a frozen feature extractor, as is common practice in many previous few-shot works~\cite{snell2017prototypical, chen2019closer, liu2020negative}.
This allows us to isolate our comparison to the method, as opposed to the learned features.
Nonetheless, the learned features have an impact on model performance.
We compare the performance of 4-layer convolutional neural networks feature extractors~\cite{vinyals2016matching} pre-trained with the ProtoNet~\cite{snell2017prototypical} and NegMargin~\cite{liu2020negative} objectives, observing $\{69.66\pm0.16, 59.88\pm0.18, 47.53\pm0.18, 35.67\pm0.17\}$ on 5-way 5-shot MiniImageNet~\cite{vinyals2016matching} with $\{0\%,20\%,40\%,60\%\}$ symmetric label swap noise.
As reported in the literature, NegMargin outperforms the ProtoNet pre-trained feature extractor when there is no support set noise during meta-test. 
On the other hand, NegMargin sees a steeper decline in performance with increasing noise levels.
We thus focus on the ProtoNet pre-trained feature extractor for our primary experiments.
We leave further investigation into this phenomenon and the performance of other feature extractor pre-training objectives on noisy few-shot learning to future work.

\minisection{Proposed ProtoNet variants: Temperature settings} As explained in Sec.~\ref{sec:sim_proto}, the temperature $T$ controls the diffuseness of the softmax for similarity weighted prototypes.
The setting of $T$ results in a trade-off between emphasizing more shots versus noise rejection capability and thus can have an impact on performance.
We show performance of similarity weighted prototypes with absolute distance, squared euclidean distance, and cosine similarity measure on MiniImageNet and TieredImageNet at varying noise levels with symmetric label swap noise, paired label swap noise, and outlier noise in Tables~\ref{tab:sym_swap_5w5k_T_sweep}, \ref{tab:pair_swap_5w5k_T_sweep}, and \ref{tab:outlier_5w5k_T_sweep}, respectively.
Note that differences in scale of $T$ for Absolute and Squared Euclidean distances versus cosine similarity is due to their scale: cosine similarity is within $\left[-1, 1\right]$, while the two distances depend on the feature dimensionality and scale.

\begin{table}[t]
\centering
\caption{
{\bf Ablation study: Binary Classification Loss} for a 3-layer TraNFS trained on 5-way 5-shot MiniImageNet~\cite{vinyals2016matching}.
}
\vspace{-3mm}
\resizebox{\columnwidth}{!}{
\begin{tabular}{c || c c c c} 
\toprule[1.2pt]
$\lambda_b$ & $0\%$ & $20\%$ & $40\%$ & $60\%$ \\
\toprule[0.8pt]		
0.0 & 68.74 $\pm$ 0.17 & 64.97 $\pm$ 0.18 & 56.29 $\pm$ 0.21 & 41.88 $\pm$ 0.23 \\
0.1 & 68.73 $\pm$ 0.17 & 65.04 $\pm$ 0.18 & 56.57 $\pm$ 0.21 & 42.23 $\pm$ 0.24 \\
0.5 & 68.53 $\pm$ 0.17 & \textbf{65.08 $\pm$ 0.18} & 56.65 $\pm$ 0.21 & \textbf{42.60 $\pm$ 0.24} \\
1.0 & 68.74 $\pm$ 0.17 & 64.81 $\pm$ 0.18 & 56.44 $\pm$ 0.21 & 42.26 $\pm$ 0.24 \\
5.0 & \textbf{68.75 $\pm$ 0.17} & 65.06 $\pm$ 0.18 & \textbf{56.71 $\pm$ 0.21} & 42.42 $\pm$ 0.24 \\
\bottomrule[1.2pt]
\end{tabular}
}
\label{tab:bin_cls_abl}
\end{table}

\begin{figure}[t]
\centering
\includegraphics[width=\columnwidth]{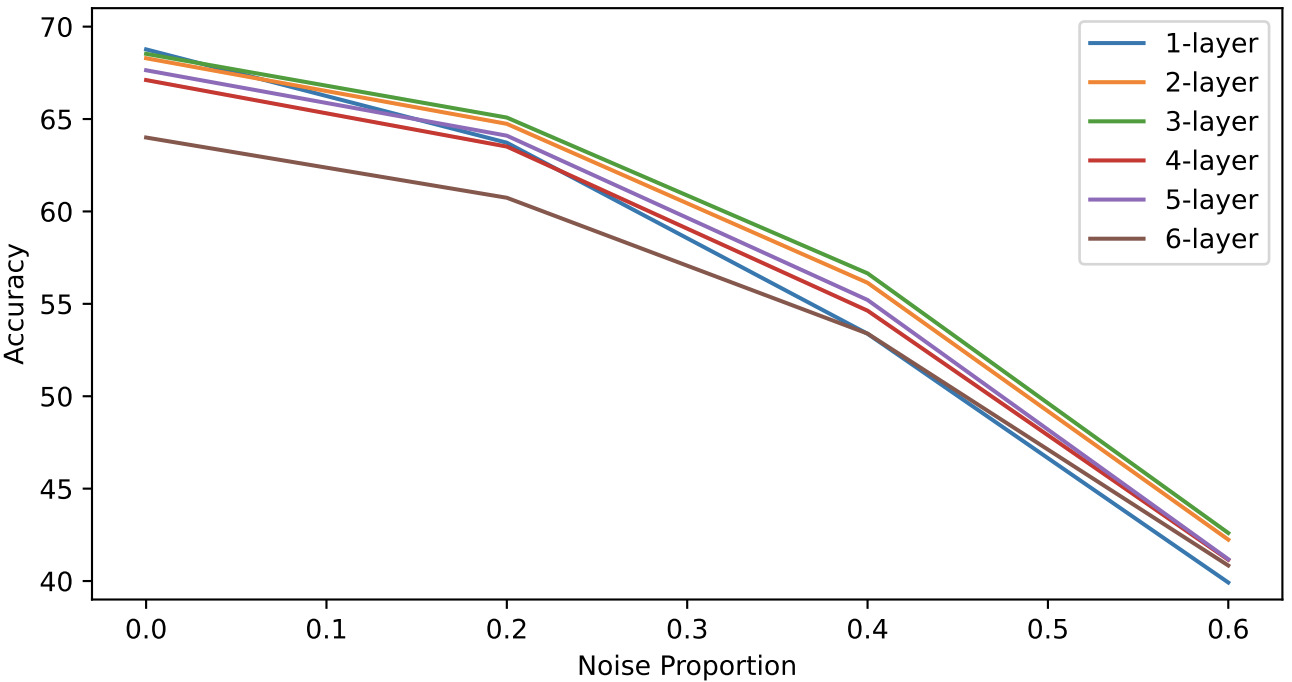}
\caption{Sweep of number of transformer layers for 5-way 5-shot MiniImageNet~\cite{vinyals2016matching} with symmetric label swap noise.}
\label{fig:layer_ablate}
\end{figure}

\begin{table*}[t!]
\centering
\caption{
{\bf Ablation study: choice of embedding for \texttt{CLS} tokens} for a 3-layer TraNFS trained on 5-way 5-shot MiniImageNet~\cite{vinyals2016matching} with symmetric label swap noise.
}
\vspace{-3mm}
\resizebox{0.7\textwidth}{!}{
\begin{tabular}{c || c c c c} 
\toprule[1.2pt]
\texttt{CLS} Token + \texttt{POS} Token & $0\%$ & $20\%$ & $40\%$ & $60\%$ \\
\toprule[0.8pt]		
Prototype + Learnable             & 68.15 $\pm$ 0.16 & 64.68 $\pm$ 0.18 & 55.04 $\pm$ 0.21 & 41.12 $\pm$ 0.22 \\
Learnable + Learnable             & 67.74 $\pm$ 0.17 & 64.28 $\pm$ 0.18 & 55.46 $\pm$ 0.22 & 41.42 $\pm$ 0.24 \\
Random Constant + Random Constant & 66.95 $\pm$ 0.17 & 63.34 $\pm$ 0.19 & 54.55 $\pm$ 0.22 & 40.87 $\pm$ 0.24 \\
Random Constant + Learnable       & \textbf{68.53 $\pm$ 0.17} & \textbf{65.08 $\pm$ 0.18} & \textbf{56.65 $\pm$ 0.21} & \textbf{42.60 $\pm$ 0.24} \\
\bottomrule[1.2pt]
\end{tabular}
}
\vspace{-2mm}
\label{tab:cls_abl}
\end{table*}

\minisection{TraNFS: Clean prototype loss}
We run a hyperparameter sweep for the loss weight term $\lambda_c$, which controls the weight of the clean prototype loss (Eq.~\eqref{eq:clean_proto}).
Results are reported in Table~\ref{tab:clean_proto_abl}.
We observe that the clean prototype loss is indeed helpful for encouraging the transformer to learn how to reject noisy samples, with a range of values of $\lambda_c$ that work well.

\minisection{TraNFS: Binary outlier detection} 
To test the effectiveness of the binary outlier classifier loss (Eq.~\eqref{eq:bin_cls}), we run a hyperparameter sweep for the loss weight term $\lambda_b$, reporting results in Table~\ref{tab:bin_cls_abl}.
We find that binary outlier classifier is indeed effective, with relatively low sensitivity to the setting of $\lambda_b$. 
Thus, we set $\lambda_b$ to be $0.5$ throughout our other experiments.

\minisection{TraNFS: \texttt{CLS} and \texttt{POS} token embeddings}
There are several options for the embeddings, corresponding to the \texttt{CLS} and \texttt{POS} tokens.
In Table~\ref{tab:cls_abl}, we meta-train a 3-layer TraNFS model on 5-shot 5-way MiniImageNet with symmetric label swap noise. Each class's \texttt{CLS} token is set using one of three options: class prototypes averaged from the convolutional embeddings, a learnable parameter, and a random constant.

While we expected the ProtoNet-style prototypes to help kick-start the transformer's comparison mechanism, we were surprised to instead observe that they underperform other choices for the \texttt{CLS} embeddings. 
After visualizing the learning curves, we observe that using prototypes as the \texttt{CLS} embeddings results in a difficult-to-escape local minimum; we hypothesize this may be the model having minimal incentive to learn anything beyond the provided prototype.
We also find that learnable \texttt{CLS} embeddings are not particularly effective: due to the random identity and shuffling of class orders between tasks, each \texttt{CLS} embedding lacks any semantic meaning beyond corresponding to a particular \texttt{POS} token's support samples; thus trying to learn some discriminitive value does not transfer between tasks and is ultimately unhelpful.
As a result, it appears that a random constant value for each \texttt{CLS} token is sufficient for the transformer.
For the \texttt{POS} positional encodings, however, learnable embeddings seem to work best. 

\minisection{TraNFS: Number of layers} 
Fig.~\ref{fig:layer_ablate} reports results for a sweep over the number of transformer layers in TraNFS. 
Matching intuition, we find that one layer is insufficient for surpassing the mean (ProtoNet) baseline. 
Different classes for each $N$-way episode mean the \texttt{CLS} embedding do not generalize across tasks.
Without prior information of what each class $c$ is in an episode, the transformer needs at least one layer to form such a concept for each position before comparisons can be made to identify samples that do not belong.
Training with too many layers, however, seems to occasionally be unstable and tends to produce slightly inferior results, perhaps due to too much overparameterization and overfitting.
We find two or three layers tend to perform best and thus report most of our results as such.
More layers in the transformer of course increases computational costs, but techniques such as knowledge distillation~\cite{hinton2015distilling, liang2021mixkd} can be used to reduce model size while minimizing performance loss.

\end{document}